\documentclass{article}
\PassOptionsToPackage{numbers, compress}{natbib}
\usepackage[final]{nips_2016}
\usepackage[utf8]{inputenc} % allow utf-8 input
\usepackage[T1]{fontenc}    % use 8-bit T1 fonts
\usepackage{url}            % simple URL typesetting
\usepackage{epsfig}
\usepackage{booktabs}       % professional-quality tables
\usepackage{amsfonts}       % blackboard math symbols
\usepackage{nicefrac}       % compact symbols for 1/2, etc.
\usepackage{microtype}      % microtypography
\usepackage{bbm}      % microtypography
\usepackage{graphicx}
\usepackage{amsmath,bm}
\usepackage{mysymbols}
\usepackage{color}
\usepackage{caption}
\usepackage[font=footnotesize,labelfont=bf]{caption}
\usepackage{times}
\usepackage{epsfig}
\usepackage{wrapfig,lipsum,booktabs}
\usepackage{graphicx}
\usepackage{amsmath}
\usepackage{amssymb}
\usepackage{csquotes}
\usepackage{algorithm}
\usepackage{algorithmic}
\usepackage[british]{babel}
\usepackage{multirow}
\usepackage{color}
\usepackage{xcolor}
\usepackage{bm}
\usepackage{xcolor}
\usepackage{subcaption}
\usepackage{booktabs}
\usepackage{enumitem}
\usepackage{lipsum}
\setlist[itemize]{leftmargin=20 pt}
\usepackage[sans]{dsfont} % just 1 with \mathds with and without sans option

\def \fsize {1pt}
\title{Hierarchical Question-Image Co-Attention \newline for Visual Question Answering}

\author{Jiasen Lu$^*$, Jianwei Yang$^*$, Dhruv Batra$^{* \dagger}$ , Devi Parikh$^{* \dagger}$\\
$*$ Virginia Tech, $\dagger$ Georgia Institute of Technology \\
{\tt\small \{jiasenlu, jw2yang, dbatra, parikh\}@vt.edu}
}
\begin{document}
\maketitle
\begin{abstract}
A number of recent works have proposed attention models for Visual Question Answering (VQA) that generate spatial maps highlighting image regions relevant to answering the question. In this paper, we argue that in addition to modeling ``where to look'' or visual attention, it is equally important to model ``what words to listen to'' or \emph{question attention}. We present a novel \emph{co-attention} model for VQA that jointly reasons about image and question attention. In addition, our model reasons about the question (and consequently the image via the co-attention mechanism) in a hierarchical fashion via a novel 1-dimensional convolution neural networks (CNN). Our model improves the state-of-the-art on the VQA dataset from 60.3\% to 60.5\%, and from 61.6\% to 63.3\% on the COCO-QA dataset. By using ResNet, the performance is further improved to 62.1\% for VQA and 65.4\% for COCO-QA.\footnote{The source code can be downloaded from \url{https://github.com/jiasenlu/HieCoAttenVQA}}.
\end{abstract}
%%%%%%%%%%%%%%%%%%%%%%%%%%%%%%%%%%%%%%%%%%%%%%%%%%%%%%%%%%%
\section{Introduction}
\label{sec:intro}
\vspace*{-2mm}
%%%%%%%%%%%%%%%%%%%%%%%%%%%%%%%%%%%%%%%%%%%%%%%%%%%%%%%%%%%
%Why VQA?  Do we need to add that general description of the paper?
Visual Question Answering (VQA) \cite{antol2015vqa, gao2015you, malinowski2015ask, ren2015exploring, zitnick2016measuring} has emerged as a prominent multi-discipline research problem in both academia and industry. To correctly answer visual questions about an image, the machine needs to understand both the image and question.
% introduce the recent progress on image attention model. Which they focus on introduce how to make the attention more "accurate". 
% should we introduce we didn't focus on this line? instead we focus on complimentary information provided in hierarchical structure and question attention. 
Recently, visual attention based models \cite{shih2015look, xiong2016dynamic, xu2015ask, yang2015stacked} have been explored for VQA, where the attention mechanism typically produces a spatial map highlighting image regions relevant to answering the question. 

%However, most (if not all) of them focus on addressing the problem of "where to look" in the image given a question, i.e., visual attention, while overlook the question side. Clearly, some words in the question are more important to focus on than others. 
So far, all attention models for VQA in literature have focused on the problem of identifying ``where to look'' or visual attention. In this paper, we argue that the problem of identifying ``which words to listen to''  or \emph{question attention} is equally important.  Consider the questions ``how many horses are in this image?'' and ``how many horses can you see in this image?". They have the same meaning, essentially captured by the first three words. A machine that attends to the first three words would arguably be more robust to linguistic variations irrelevant to the meaning and answer of the question. Motivated by this observation, in addition to reasoning about visual attention, we also address the problem of question attention. Specifically, we present a novel multi-modal attention model for VQA with the following two unique features:

\textbf{Co-Attention}: We propose a novel mechanism that jointly reasons about visual attention and question attention, which we refer to as \emph{co-attention}. Unlike previous works, which only focus on visual attention, our model has a natural symmetry between the image and question, in the sense that the image representation is used to guide the question attention and the question representation(s) are used to guide image attention. 

%Joint attention has been seen in some previous literatures in the natural language processing area. To the best of our knowledge, we are the first to apply co-attention in such a cross-modality application as VQA.

%Upon the visual attention strategy, some previous works performed attention for multiple times in a recursive (stacked) manner \cite{xiong2016dynamic, yang2015stacked}. Another kind of strategy called multi-hop attention is recently proposed in \cite{xu2015ask}. It aligns words to image patches at the first hop, and then consider the whole question to obtain improved result at the second hop. In this process, the attention at first hop is used as the cue for attention at the second hop. Inspired by this work, 
\textbf{Question Hierarchy}: We build a hierarchical architecture that co-attends to the image and question at three levels: (a) word level, (b) phrase level and (c) question level. At the word level, we embed the words to a vector space through an embedding matrix. At the phrase level, 1-dimensional convolution neural networks are used to capture the information contained in unigrams, bigrams and trigrams. Specifically, we convolve word representations with temporal filters of varying support, and then combine the various n-gram responses by pooling them into a single phrase level representation. At the question level, we use recurrent neural networks to encode the entire question. For each level of the question representation in this hierarchy, we construct joint question and image co-attention maps, which are then combined recursively to ultimately predict a distribution over the answers. 

Overall, the main contributions of our work are:
\begin{itemize}
\item We propose a novel co-attention mechanism for VQA that jointly performs question-guided visual attention and image-guided question attention. We explore this mechanism with two strategies, parallel and alternating co-attention, which are described in Sec.~\ref{subsec:co-attention};
\item We propose a hierarchical architecture to represent the question, and consequently construct image-question co-attention maps at 3 different levels: word level, phrase level and question level. These co-attended features are then recursively combined from word level to question level for the final answer prediction;
\item At the phrase level, we propose a novel convolution-pooling strategy to adaptively select the phrase sizes whose representations are passed to the question level representation;
\item Finally, we evaluate our proposed model on two large datasets, VQA \cite{antol2015vqa} and COCO-QA \cite{ren2015exploring}. We also perform ablation studies to quantify the roles of different components in our model.
\end{itemize}
%in this paper, we propose an end-to-end hierarchical image and question attention networks, that allow multi-level reasoning between question and image. More specifically, we consturct word level, phrase level and sentence level question and image attentions by using two kinds of interactive attetion methods. From word level to phrase level, we use max pooling across n gram representations at each time step. From phrase level to sentence level, we use LSTM to encode the phrase representation at a sequential order. Given the attention features at each stage, we use a LSTM to reason the attention transition and predict the answer. We test our proposed networks on VQA and COCO-QA dataset and improve the state-of-the-art performance. 
\begin{figure}[t]
 \centering
 \includegraphics[width=0.86\linewidth, scale=1]{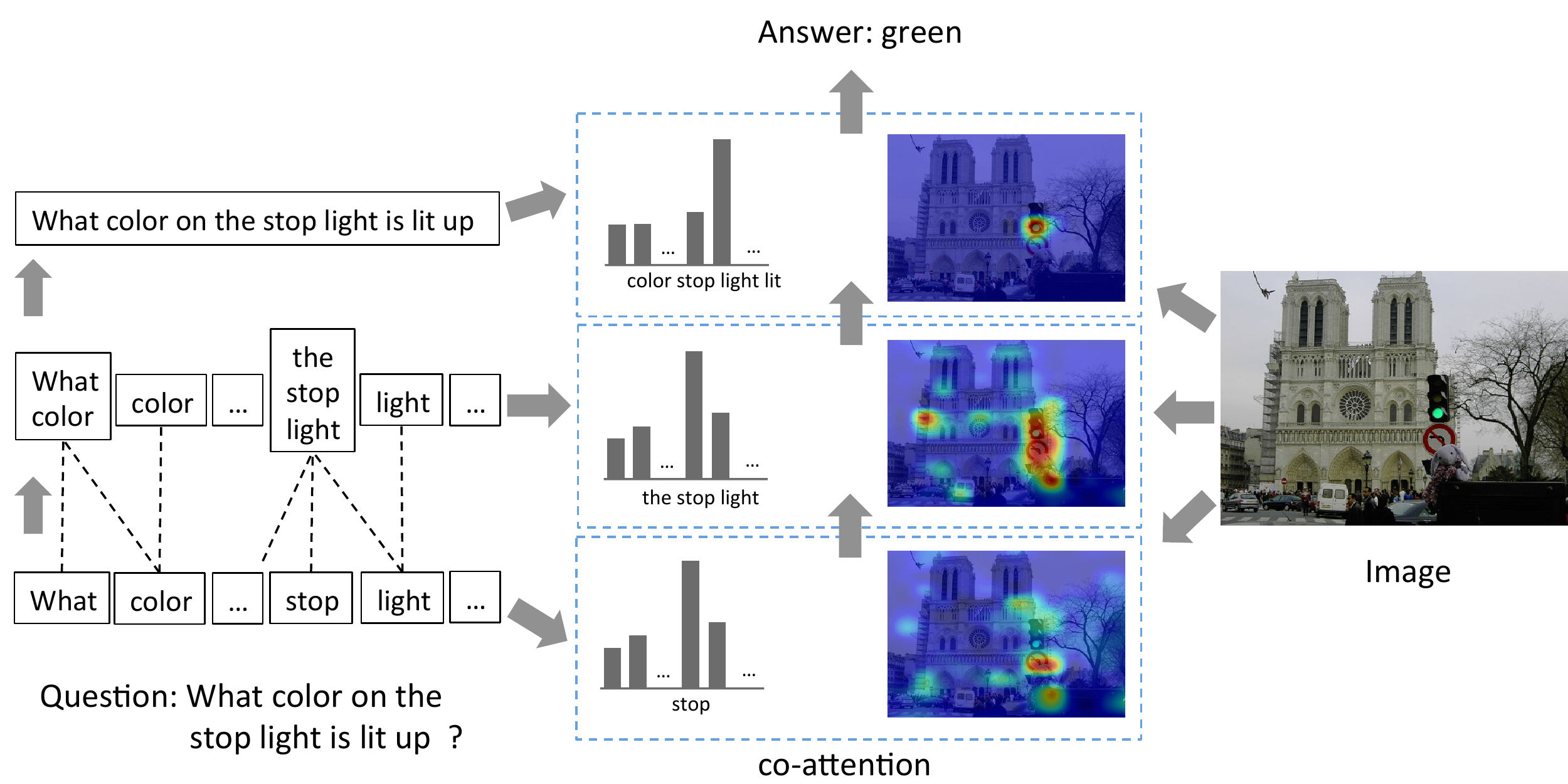}
 \caption{Flowchart of our proposed hierarchical co-attention model. Given a question, we extract its word level, phrase level and question level embeddings. At each level, we apply co-attention on both the image and question. The final answer prediction is based on all the co-attended image and question features.}
 \label{fig:teaser}
\end{figure}

%%%%%%%%%%%%%%%%%%%%%%%%%%%%%%%%%%%%%%%%%%%%%%%%%%%%%%%%%%%
\section{Related Work}
\label{sec:related}
%%%%%%%%%%%%%%%%%%%%%%%%%%%%%%%%%%%%%%%%%%%%%%%%%%%%%%%%%%%
%A number of 
Many recent works \cite{antol2015vqa, gao2015you, krishna2016visual, malinowski2015ask, ren2015exploring, zhang2015yin, kim2016multimodal, fukui2016multimodal} have proposed models for VQA. We compare and relate our proposed co-attention mechanism to other vision and language attention mechanisms in literature. 

%Many recent works on VQA have employed attention mechanism for region focus.
\textbf{Image attention}. Instead of directly using the holistic entire-image embedding from the fully connected layer of a deep CNN (as in \cite{antol2015vqa, ma2015learning, malinowski2015ask, ren2015exploring}), a number of recent works have explored image attention models for VQA. Zhu \etal \cite{zhu2015visual7w} add spatial attention to the standard LSTM model for pointing and grounded QA. Andreas \etal \cite{andreas2015deep} propose a compositional scheme that consists of a language parser and a number of neural modules networks. The language parser predicts which neural module network should be instantiated to answer the question. 
%One of these neural modules networks involves attending to specific regions in an image. 
Some other works perform image attention multiple times in a stacked manner. In \cite{yang2015stacked}, the authors propose a stacked attention network, which runs multiple hops to infer the answer progressively. To capture fine-grained information from the question, Xu \etal \cite{xu2015ask} propose a multi-hop image attention scheme. It aligns words to image patches in the first hop, and then refers to the entire question for obtaining image attention maps in the second hop. In \cite{shih2015look}, the authors generate image regions with object proposals and then select the regions relevant to the question and answer choice. Xiong~\etal \cite{xiong2016dynamic} augments dynamic memory network with a new input fusion module and retrieves an answer from an attention based GRU. In concurrent work, \cite{das2016human} collected `human attention maps' that are used to evaluate the attention maps generated by attention models for VQA. Note that all of these approaches model visual attention alone, and do not model question attention. Moreover, \cite{xu2015ask, yang2015stacked} model attention sequentially, i.e., later attention is based on earlier attention, which is prone to error propagation. In contrast, we conduct co-attention at three levels independently.

%Ilievski et al (2016) who run state-of-the-art object detector of the classes extracted from the key words in the question. In contrast to other attention mechanisms, such approach offers a focused, question dependent, “hard” attention.
\textbf{Language Attention}. Though no prior work has explored question attention in VQA, there are some related works in natural language processing (NLP) in general that have modeled language attention. In order to overcome difficulty in translation of long sentences, Bahdanau \etal \cite{bahdanau2014neural} propose RNNSearch to learn an alignment over the input sentences. In \cite{hermann2015teaching}, the authors propose an attention model to circumvent the bottleneck caused by fixed width hidden vector in text reading and comprehension. A more fine-grained attention mechanism is proposed in \cite{rocktaschel2015reasoning}. The authors employ a word-by-word neural attention mechanism to reason about the entailment in two sentences. Also focused on modeling sentence pairs, the authors in \cite{yin2015abcnn} propose an attention-based bigram CNN for jointly performing attention between two CNN hierarchies. In their work, three attention schemes are proposed and evaluated. In \cite{santos2016attentive}, the authors propose a two-way attention mechanism to project the paired inputs into a common representation space.
\section{Method}
\label{sec:method}
\vspace*{-2mm}
We begin by introducing the notation used in this paper. To ease understanding, our full model is described in parts. First, our hierarchical question representation is described in Sec.~\ref{subsec:hierarchy} and the proposed co-attention mechanism is then described in Sec.~\ref{subsec:co-attention}. Finally, Sec.~\ref{subsec:recursive} shows how to recursively combine the attended question and image features to output answers.
%%%%%%%%%%%%%%%%%%%%%%%%%%%%%%%%%%%%%%%%%%%%%%%%%%%%%%%%%%%
\subsection{Notation}
Given a question with $T$ words, its representation is denoted by $\bm{Q}=\left\lbrace \bm{q}_1,\ldots \bm{q}_T \right\rbrace$, where $\bm{q}_t$ is the feature vector for the $t$-th word. We denote $\bm{q}^w_t$, $\bm{q}^p_t$ and $\bm{q}^s_t$ as word embedding, phrase embedding and question embedding at position $t$, respectively. 
%For an image, the response of the last convolutional layer of VGGNet \cite{Simonyan14c} or Deep Residual network \cite{he2015deep} are used, and 
The image feature is denoted by $\bm{{V}} = \{\bm{v}_1,...,\bm{v}_N\}$, where $\bm{v}_n$ is the feature vector at the spatial location $n$. The co-attention features of image and question at each level in the hierarchy are denoted as $\hat{\bm{v}}^r$ and $\hat{\bm{q}}^r$ where $r \in \left\lbrace w, p, s \right\rbrace $. The weights in different modules/layers are denoted with $\bm{W}$, with appropriate sub/super-scripts as necessary. In the exposition that follows, we omit the bias term $\bm{b}$ to avoid notational clutter.
\subsection{Question Hierarchy}
\label{subsec:hierarchy}
Given the 1-hot encoding of the question words $\bm{Q}=\left\lbrace \bm{q}_1,\ldots, \bm{q}_T \right\rbrace$, we first embed the words to a vector space (learnt end-to-end) to get  $\bm{Q}^w = \left\lbrace \bm{q}^w_1,\ldots, \bm{q}^w_T \right\rbrace$. To compute the phrase features, we apply 1-D convolution on the word embedding vectors. Concretely, at each word location, we compute the inner product of the word vectors with filters of three window sizes: unigram, bigram and trigram. For the $t$-th word, the convolution output with window size $s$ is given by
\begin{equation}
\hat{\bm{q}}^p_{s,t}=\tanh(\bm{W}_c^s \bm{q}^w_{t:t+s-1}), \quad s\in \{1,2,3\}
\end{equation}
where $\bm{W}_c^s$ is the weight parameters. The word-level features $\bm{Q}^w$  are appropriately 0-padded before feeding into bigram and trigram convolutions to maintain the length of the sequence after convolution. Given the convolution result, we then apply max-pooling across different n-grams at each word location to obtain phrase-level features
\begin{equation}
\bm{q}^p_{t} = \max (\hat{\bm{q}}^p_{1,t}, \hat{\bm{q}}^p_{2,t}, \hat{\bm{q}}^p_{3,t}), \quad t \in \{1, 2,\ldots,T \}
\end{equation}

Our pooling method differs from those used in previous works \cite{hu2014convolutional} in that it adaptively selects different gram features at each time step, while preserving the original sequence length and order.
We use a LSTM to encode the sequence $\bm{q}^p_t$ after max-pooling. The corresponding question-level feature $\bm{q}^s_{t}$ is the LSTM hidden vector at time $t$. 

Our hierarchical representation of the question is depicted in Fig.~\ref{fig:added}(a).
%comining the outputs of two directions, i.e., $\bm{q}^s_t = \overleftarrow{\bm{q}}^s_t \Vert \overrightarrow{\bm{q}}^s_t$.
\subsection{Co-Attention}
\label{subsec:co-attention}
\begin{figure}[t]
 \centering 
 \includegraphics[width=0.87\linewidth, scale=1]{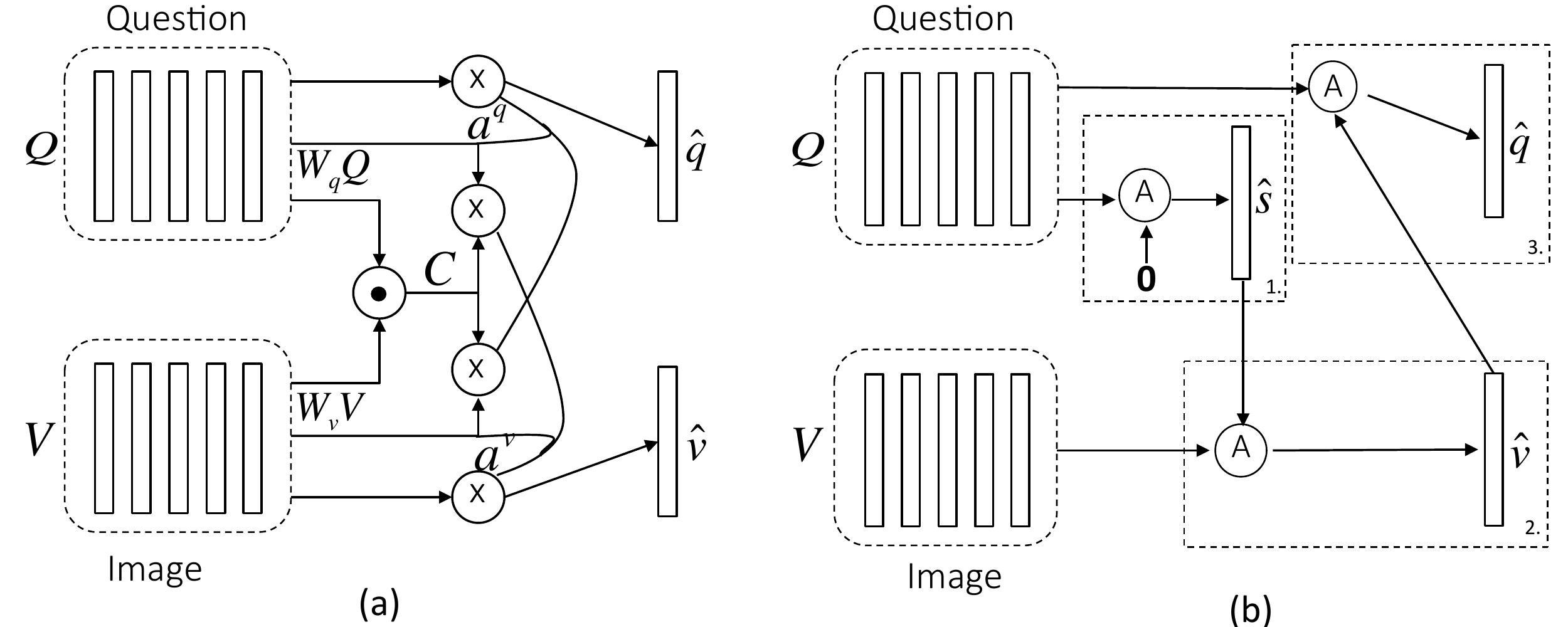}
 \caption{(a) Parallel co-attention mechanism; (b) Alternating co-attention mechanism.}
 \label{fig:model}
\end{figure}

We propose two co-attention mechanisms that differ in the order in which image and question attention maps are generated. The first mechanism, which we call parallel co-attention, generates image and question attention simultaneously. The second mechanism, which we call alternating co-attention, sequentially alternates between generating image and question attentions. See Fig.~\ref{fig:model}. These co-attention mechanisms are executed at all three levels of the question hierarchy.

\textbf{Parallel Co-Attention}. Parallel co-attention attends to the image and question simultaneously. Similar to \cite{xu2015ask}, we connect the image and question by calculating the similarity between image and question features at all pairs of image-locations and question-locations. Specifically, given an image feature map $\bm{{V}} \in \mathcal{R}^{d \times N}$, and the question representation $\bm{{Q}} \in \mathcal{R}^{d \times T}$, the affinity matrix $\bm{C} \in \mathcal{R}^{T \times N}$ is calculated by
\begin{equation}
\bm{C} = \tanh(\bm{Q}^T \bm{W}_{b} \bm{V})
\end{equation}
where $\bm{W}_b \in \mathcal{R}^{d \times d}$ contains the weights. After computing this affinity matrix, one possible way of computing the image (or question) attention is to simply maximize out the affinity over the locations of other modality, \ie $\bm{a}^v[n] = \max_i(\bm{C}_{i,n}) $ and $\bm{a}^q[t] = \max_j(\bm{C}_{t,j})$. 
Instead of choosing the max activation, we find that performance is improved if we consider this affinity matrix as a feature and learn to predict image and question attention maps via the following

%\begin{equation}
%\bm{a}^v = \textrm{softmax}(\bm{w}^T_{cv} \bm{C}), \quad \bm{a}^q=\textrm{softmax}(\bm{w}^T_{cq} \bm{C} ^T)\end{equation}
%\end{equation}
\begin{equation}
\begin{aligned}
\bm{H}^v = \tanh(\bm{W}_v \bm{V} + (\bm{W}_{q} \bm{Q})\bm{C}) , &\quad \bm{H}^q = \tanh(\bm{W}_q \bm{Q} + (\bm{W}_{v} \bm{V}) \bm{C}^T) \\ 
\bm{a}^v = \textrm{softmax}(\bm{w}_{hv}^T \bm{H}^v) , &\quad \bm{a}^q = \textrm{softmax}(\bm{w}_{hq}^T \bm{H}^q)\\
\end{aligned}
\end{equation}
where $\bm{W}_v, \bm{W}_q \in \mathcal{R}^{k \times d}$, $\bm{w}_{hv}, \bm{w}_{hq} \in \mathcal{R}^k$ are the weight parameters. $\bm{a}^v \in \mathcal{R}^N$ and $\bm{a}^q \in \mathcal{R}^T$ are the attention probabilities of each image region $\bm{v}_n$ and word $\bm{q}_t$ respectively. The affinity matrix $\bm{C}$ transforms question attention space to image attention space (vice versa for $\bm{C}^T$).
%\textcolor{red}{where $W_v$, $W_q$ are the blah blah, $H^v$ is the blah blah.} 
Based on the above attention weights, the image and question attention vectors are calculated as the weighted sum of the image features and question features, i.e.,
\begin{equation}
\hat{\bm{v}} = \sum_{n = 1}^{N} a^v_n \bm{v}_n,  \quad  \hat{\bm{q}} = \sum_{t=1}^{T} a^q_t \bm{q}_t
\end{equation}
The parallel co-attention is done at each level in the hierarchy, leading to $\hat{\bm{v}}^{r}$ and $\hat{\bm{q}}^{r}$ where $r \in \left\lbrace w, p, s \right\rbrace$.
%One can aslo use the attention GRU proposed in \cite{xiong2016dynamic}, which can preserve the location and order information compare to soft attention. For simplicity, we only use soft attention in our paper.

\textbf{Alternating Co-Attention.} In this attention mechanism, we sequentially alternate between generating image and question attention. Briefly, this consists of three steps (marked in Fig.~\ref{fig:model}b): 1) summarize the question into a single vector $\bm{q}$; 2) attend to the image based on the question summary $\bm{q}$; 3) attend to the question based on the attended image feature. 

Concretely, we define an attention operation $\hat{\bm{x}} = \mathcal{A}({\bm{X}}; \bm{g})$, which takes the image (or question) features $\bm{X}$ and attention guidance $\bm{g}$ derived from question (or image) as inputs, and outputs the attended image (or question) vector. The operation can be expressed in the following steps
\begin{equation}
\begin{aligned}
\bm{H} &= \tanh(\bm{W}_x \bm{X} + (\bm{W}_{g} \bm{g}) \bm{\mathds{1}}^T) \\ 
\bm{a}^x &= \textrm{softmax}(\bm{w}_{hx}^T \bm{H}) \\
\hat{\bm{x}} &= \sum a^x_i \bm{x}_i
\end{aligned}
\end{equation}
where $\bm{\mathds{1}}$ is a vector with all elements to be 1. $\bm{W}_x, \bm{W}_g \in \mathcal{R}^{k \times d}$ and $\bm{w}_{hx} \in \mathcal{R}^k$ are parameters. $\bm{a}^x$ is the attention weight of feature $\bm{X}$. 

The alternating co-attention process is illustrated in Fig. \ref{fig:model} (b). At the first step of alternating co-attention, $\bm{X} = \bm{Q}$, and $\bm{g}$ is $\bm{0}$; At the second step, $\bm{X} = \bm{V}$ where $\bm{V}$ is the image features, and the guidance $\bm{g}$ is intermediate attended question feature $\hat{\bm{s}}$ from the first step; Finally, we use the attended image feature $\hat{\bm{v}}$ as the guidance to attend the question again, i.e., $\bm{X} = \bm{{Q}}$ and $\bm{g} = \hat{\bm{v}}$. Similar to the parallel co-attention, the alternating co-attention is also done at each level of the hierarchy.
\subsection{Encoding for Predicting Answers}
\label{subsec:recursive}

\begin{figure}[t]
 \centering 
 \includegraphics[width=0.87\linewidth, scale=1]{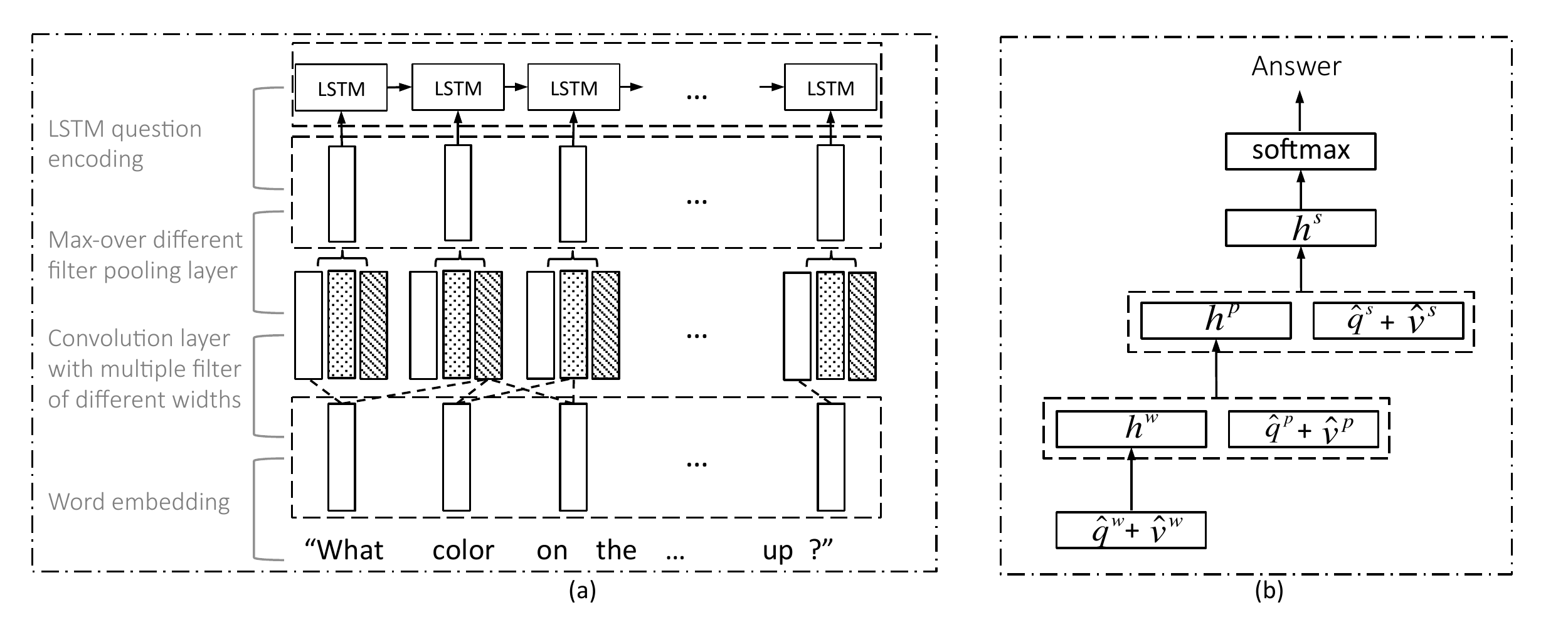}
 \caption{(a) Hierarchical question encoding (Sec.~\ref{subsec:hierarchy}); (b) Encoding for predicting answers (Sec.~\ref{subsec:recursive}).}
 \label{fig:added}
\end{figure}

Following \cite{antol2015vqa}, we treat VQA as a classification task. We predict the answer based on the co-attended image and question features from all three levels. We use a multi-layer perceptron (MLP) to recursively encode the attention features as shown in Fig.~\ref{fig:added}(b).
\begin{equation}
\begin{aligned}
\bm{h}^w &= \tanh(\bm{W}_{w} (\bm{\hat{q}^w} + \bm{\hat{v}^w})) \\
\bm{h}^p &= \tanh(\bm{W}_{p} [(\bm{\hat{q}^p} + \bm{\hat{v}^p}), \bm{h}^w]) \\
\bm{h}^s &= \tanh(\bm{W}_{s} [(\bm{\hat{q}^s} + \bm{\hat{v}^s}), \bm{h}^p]) \\
\bm{p} &= \textrm{softmax}(\bm{W}_{h} \bm{h}^s) \\
\end{aligned}
\end{equation}
where $\bm{W}_{w}, \bm{W}_{p}, \bm{W}_{s}$ and $\bm{W}_{h}$ are the weight parameters. $[\cdot]$ is the concatenation operation on two vectors. $\bm{p}$ is the probability of the final answer.  
%Following $\bm{h}$, we append a softmax layer for predicting the probability of each answer. 
%%%%%%%%%%%%%%%%%%%%%%%%%%%%%%%%%%%%%%%%%%%%%%%%%%%%%%%%%%%
\section{Experiment}
\vspace*{-2mm}
\label{sec:exp}
%%%%%%%%%%%%%%%%%%%%%%%%%%%%%%%%%%%%%%%%%%%%%%%%%%%%%%%%%%%
\subsection{Datasets}
We evaluate the proposed model on two datasets, the VQA dataset \cite{antol2015vqa} and the COCO-QA dataset \cite{ren2015exploring}.

\textbf{VQA} dataset \cite{antol2015vqa} is the largest dataset for this problem, containing human annotated questions and answers on Microsoft COCO dataset \cite{lin2014microsoft}. The dataset contains 248,349 training questions, 121,512 validation questions, 244,302 testing questions, and a total of 6,141,630 question-answers pairs. There are three sub-categories according to answer-types including yes/no, number, and other. Each question has 10 free-response answers. We use the top 1000 most frequent answers as the possible outputs similar to \cite{antol2015vqa}. This set of answers covers 86.54\% of the train+val answers. For testing, we train our model on VQA train+val and report the test-dev and test-standard results from the VQA evaluation server. We use the evaluation protocol of \cite{antol2015vqa} in the experiment.

\textbf{COCO-QA} dataset \cite{ren2015exploring} is automatically generated from captions in the Microsoft COCO dataset \cite{lin2014microsoft}. There are 78,736 train questions and 38,948 test questions in the dataset. These questions are based on 8,000 and 4,000 images respectively.  There are four types of questions including object, number, color, and location. Each type takes $70\%$, $7\%$, $17\%$, and $6\%$ of the whole dataset, respectively. All answers in this data set are single word. As in \cite{ren2015exploring}, we report classification accuracy as well as Wu-Palmer similarity (WUPS) in Table 2.
\subsection{Setup}
We use Torch \cite{torch} to develop our model. We use the Rmsprop optimizer with a base learning rate of 4e-4, momentum 0.99 and weight-decay 1e-8. We set batch size to be 300 and train for up to 256 epochs with early stopping if the validation accuracy has not improved in the last 5 epochs. For COCO-QA, the size of hidden layer $\bm{W_s}$ is set to 512 and 1024 for VQA since it is a much larger dataset. All the other word embedding and hidden layers were vectors of size 512. We apply dropout with probability $0.5$ on each layer. Following \cite{yang2015stacked}, we rescale the image to $448 \times 448$, and then take the activation from the last pooling layer of VGGNet \cite{Simonyan14c} or ResNet \cite{he2015deep} as its feature.
%Our model is end-to-end trainable, but we do not finetune the CNN. 
%The code is available at \url{https://github.com/jiasenlu/HieCoAttenVQA}.
%
\subsection{Results and Analysis}
There are two test scenarios on VQA: open-ended and multiple-choice. The best performing method \textbf{deeper LSTM Q + norm I} from \cite{antol2015vqa} is used as our baseline. For open-ended test scenario, we compare our method with the recent proposed \textbf{SMem} \cite{xu2015ask}, \textbf{SAN} \cite{yang2015stacked}, \textbf{FDA} \cite{Ilievski2016} and \textbf{DMN+} \cite{xiong2016dynamic}. For multiple choice, we compare with \textbf{Region Sel.} \cite{shih2015look} and \textbf{FDA} \cite{Ilievski2016}. We compare with \textbf{2-VIS+BLSTM} \cite{ren2015exploring}, \textbf{IMG-CNN} \cite{ma2015learning} and \textbf{SAN} \cite{yang2015stacked} on COCO-QA. We use $\textrm{Ours}^p$ to refer to our parallel co-attention, $\textrm{Ours}^a$ for alternating co-attention.

Table~\ref{tab:vqa} shows results on the VQA test sets for both open-ended and multiple-choice settings. We can see that our approach improves the state of art from 60.4\% (DMN+ \cite{xiong2016dynamic}) to 62.1\% ($\textrm{Ours}^a$+ResNet) on open-ended and from 64.2\% (FDA \cite{Ilievski2016}) to 66.1\% ($\textrm{Ours}^a$+ResNet) on multiple-choice. Notably, for the question type \textit{Other} and \textit{Num}, we achieve 3.4\% and 1.4\% improvement on open-ended questions, and 4.0\% and 1.1\% on multiple-choice questions. As we can see, ResNet features outperform or match VGG features in all cases. Our improvements are not solely due to the use of a better CNN. Specifically, FDA \cite{Ilievski2016} also uses ResNet \cite{he2015deep}, but $\textrm{Ours}^a$+ResNet outperforms it by 1.8\% on test-dev. SMem \cite{xu2015ask} uses GoogLeNet \cite{szegedy2015going} and the rest all use VGGNet \cite{Simonyan14c}, and Ours+VGG outperforms them by 0.2\% on test-dev (DMN+ \cite{xiong2016dynamic}). 

Table~\ref{tab:cocoqa} shows results on the COCO-QA test set. Similar to the result on VQA, our model improves the state-of-the-art from 61.6\% (SAN(2,CNN)~\cite{yang2015stacked}) to 65.4\% ($\textrm{Ours}^a$+ResNet). We observe that parallel co-attention performs better than alternating co-attention in this setup. Both attention mechanisms have their advantages and disadvantages: parallel co-attention is harder to train because of the dot product between image and text which compresses two vectors into a single value. On the other hand, alternating co-attention may suffer from errors being accumulated at each round. %Due to significant computational requirements, we could not train $Ours^p$+Residual on VQA and COCO-QA. This is left as future work.
%Combining the two attention mechanisms into a single model is fairly straightforward, but left for future work largely due to computational constraints.
%As a result, we suspect that combining two co-attention strategies may further improve the performance, though we do not conduct such experiments for limited computational resources.
%
\begin{table}[t]\footnotesize
\setlength{\tabcolsep}{5.5pt}
  \caption{Results on the VQA dataset. ``-'' indicates the results is not available.}% \textcolor{red}{``$^\ast$'' indicates finetune on CNN.}}
  \centering
  \begin{tabular}{l c c c c c c c c c c}
    \toprule
    \multicolumn{1}{c}{} & \multicolumn{5}{c}{Open-Ended}  & \multicolumn{5}{c}{Multiple-Choice}\\
   \cmidrule(r){2-6}
   \cmidrule(r){7-11}
    \multicolumn{1}{c}{}   & \multicolumn{4}{c}{test-dev}  & \multicolumn{1}{c}{test-std}   & \multicolumn{4}{c}{test-dev}  & \multicolumn{1}{c}{test-std} \\
    \cmidrule(r){2-5}
    \cmidrule(r){6-6}    
    \cmidrule(r){7-10}
    \cmidrule(r){11-11}
    Method  & Y/N  & Num & Other & All & All & Y/N  & Num & Other & All & All\\
    \midrule	    
    LSTM Q+I \cite{antol2015vqa}& 80.5 & 36.8 & 43.0 & 57.8 & 58.2 & 80.5 & 38.2 & 53.0 & 62.7 & 63.1  \\
	Region Sel. \cite{shih2015look} & - & - & - & - & - & 77.6 & 34.3 & 55.8 & 62.4& - \\   
    SMem \cite{xu2015ask} & 80.9 &  37.3  & 43.1  & 58.0 & 58.2 & - & - & - & - & - \\
    SAN \cite{yang2015stacked}  & 79.3 &  36.6  & 46.1   &  58.7 & 58.9 & - & - & - & - & - \\
    FDA \cite{Ilievski2016}  & \textbf{81.1} &  36.2  & 45.8   &  59.2 & 59.5 & \textbf{81.5} &  39.0 & 54.7 & 64.0 & 64.2 \\
    DMN+ \cite{xiong2016dynamic} & 80.5 &  36.8 &  48.3  & 60.3  & 60.4 & - & - & - & - & - \\
    \midrule	    
    $\textrm{Ours}^p$+VGG & 79.5 & \textbf{38.7} & 48.3 & 60.1 & - & 79.5 & 39.8 & 57.4 & 64.6 &-\\
    $\textrm{Ours}^a$+VGG & 79.6 & 38.4 & 49.1 & 60.5 & - & 79.7 & \textbf{40.1} & 57.9 & 64.9 &-\\    
    $\textrm{Ours}^a$+ResNet & 79.7 &  \textbf{38.7} & \textbf{51.7} & \textbf{61.8} & \textbf{62.1} & 79.7 & 40.0 & \textbf{59.8} & \textbf{65.8} & \textbf{66.1}\\
    \bottomrule
  \end{tabular}
  \label{tab:vqa}
\end{table}  
\begin{table}[t]\footnotesize
\setlength{\tabcolsep}{5.5pt}
  \caption{Results on the COCO-QA dataset. ``-'' indicates the results is not available.} %\textcolor{red}{``$^\ast$'' indicates finetune on CNN.}}
  \label{tab:cocoqa}
  \centering
  \begin{tabular}{l c c c c c c c}
    \toprule
    Method  & Object & Number & Color & Location & Accuracy & WUPS0.9 & WUPS0.0 \\
    \midrule	    
	2-VIS+BLSTM \cite{ren2015exploring} & 58.2 & 44.8 & 49.5 & 47.3 & 55.1 & 65.3 & 88.6\\
	IMG-CNN \cite{ma2015learning} & - & - & - & - & 58.4 &  68.5 &  89.7\\	
	SAN(2, CNN) \cite{yang2015stacked} & 64.5 & 48.6 & 57.9 & 54.0 & 61.6 & 71.6 &  90.9 \\
    \midrule	    	 
	 $\textrm{Ours}^p$+VGG & 65.6 & 49.6 & 61.5 & 56.8 & 63.3 & 73.0 & 91.3\\	
	 $\textrm{Ours}^a$+VGG & 65.6 & 48.9 & 59.8 & 56.7  & 62.9 & 72.8 & 91.3\\
	 $\textrm{Ours}^a$+ResNet & \textbf{68.0} & \textbf{51.0} & \textbf{62.9} & \textbf{58.8} & \textbf{65.4}& \textbf{75.1} & \textbf{92.0} \\
   \bottomrule
  \end{tabular}
\end{table}
%\textcolor{red}{this subsection qualitively show our system is more robust on question paraphrase. We also show the corresponding attention maps}
%
\subsection{Ablation Study}
\label{subsec:ablation}
In this section, we perform ablation studies to quantify the role of each component in our model. Specifically, we re-train our approach by ablating certain components:
\begin{itemize}
\item \emph{Image Attention alone}, where in a manner similar to previous works \cite{yang2015stacked}, we do not use any question attention. The goal of this comparison is to verify that our improvements are not the result of orthogonal contributions. (say better optimization or better CNN features). 
\item \emph{Question Attention alone}, where no image attention is performed.
\item \emph{W/O Conv}, where no convolution and pooling is performed to represent phrases. Instead, we stack another word embedding layer on the top of word level outputs. 
%The goal of this comparison is to verify whether our 1D convolution+pooling strategy over words indeed better captures local information from question fragments. 
\item \emph{W/O W-Atten}, where no word level co-attention is performed. We replace the word level attention with a uniform distribution. Phrase and question level co-attentions are still modeled.
\item \emph{W/O P-Atten}, where no phrase level co-attention is performed, and the phrase level attention is set to be uniform. Word and question level co-attentions are still modeled.
\item \emph{W/O Q-Atten}, where no question level co-attention is performed. We replace the question level attention with a uniform distribution. Word and phrase level co-attentions are still modeled.
\end{itemize}

Table~\ref{tab:ablation} shows the comparison of our full approach w.r.t these ablations on the VQA validation set (test sets are not recommended to be used for such experiments). The \textbf{deeper LSTM Q + norm I} baseline in \citep{antol2015vqa} is also reported for comparison. We can see that image-attention-alone does improve performance over the holistic image feature (\textbf{deeper LSTM Q + norm I}), which is consistent with findings of previous attention models for VQA \cite{xiong2016dynamic, yang2015stacked}.

\begin{wraptable}{r}{6.5cm}\footnotesize
    \centering
    \captionof{table}{Ablation study on the VQA dataset using $\textrm{Ours}^a$+VGG.}
    \label{tab:ablation}
    \begin{tabular}{l c c c c}
    \toprule
    \multicolumn{1}{c}{}   & \multicolumn{4}{c}{validation} \\
    \cmidrule{2-5}
    Method  & Y/N  & Num & Other & All \\
    \midrule	    
    LSTM Q+I & \textbf{79.8} & 32.9 & 40.7 & 54.3 \\
	Image Atten & 79.8& 33.9& 43.6& 55.9\\   	
	Question Atten & 79.4& 33.3& 41.7& 54.8\\
    W/O Q-Atten &79.6 &  32.1&  42.9&  55.3\\ 
    W/O P-Atten & 79.5& 34.1& 45.4& 56.7\\
    W/O W-Atten & 79.6& 34.4& 45.6& 56.8\\
   	Full Model & 79.6 &  \textbf{35.0} & \textbf{45.7} & \textbf{57.0}\\
    \bottomrule
    \end{tabular}       
\end{wraptable}
Comparing the full model \wrt ablated versions without {word, phrase, question} level attentions reveals a clear interesting trend -- the attention mechanisms closest to the `top' of the hierarchy (\ie question) matter most, with a drop of 1.7\% in accuracy if not modeled; followed by the intermediate level (\ie phrase), with a drop of 0.3\%; finally followed by the `bottom' of the hierarchy (\ie word), with a drop of 0.2\% in accuracy. We hypothesize that this is because the question level is the `closest' to the answer prediction layers in our model. Note that \emph{all} levels are important, and our final model significantly outperforms not using any linguistic attention (1.1\% difference between Full Model and Image Atten). The question attention alone model is better than LSTM Q+I, with an improvement of 0.5\% and worse than image attention alone, with a drop of 1.1\%. $\textrm{Ours}^a$ further improves if we performed alternating co-attention for one more round, with an improvement of 0.3\%.
\subsection{Qualitative Results}
\begin{figure}[tb]
\center
\begin{minipage}{0.22\linewidth}
\includegraphics[width=1\linewidth,  height=1\linewidth, scale=0.8, trim={0 0in 0 0in}, clip]{demo/coco_qa/img_0.jpg}
\end{minipage}
\begin{minipage}{0.22\linewidth}
\includegraphics[width=1\linewidth,  height=1\linewidth, scale=0.8, trim={0 0in 0 0in}, clip]{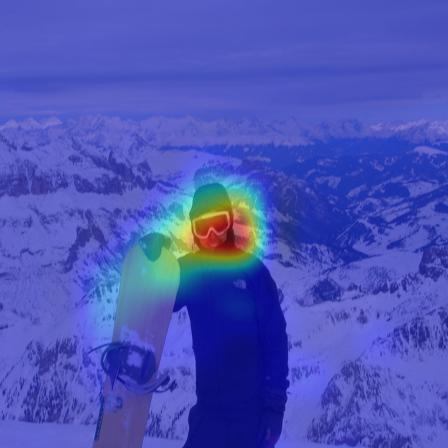}
\end{minipage}
\begin{minipage}{0.22\linewidth}
\includegraphics[width=1\linewidth,  height=1\linewidth, scale=0.8, trim={0 0in 0 0in}, clip]{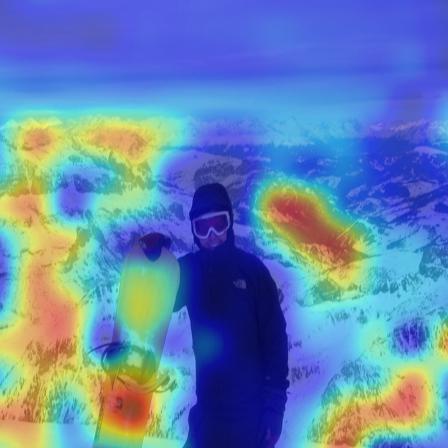}
\end{minipage}
\begin{minipage}{0.22\linewidth}
\includegraphics[width=1\linewidth,  height=1\linewidth, scale=0.8, trim={0 0in 0 0in}, clip]{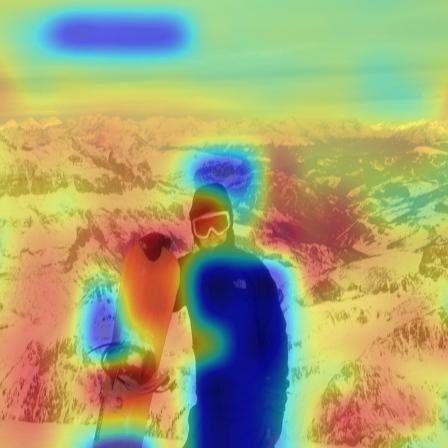}
\end{minipage}
\\
\begin{minipage}{0.22\linewidth}
\begin{center}
\tiny{\textbf{Q}: what is the man holding a snowboard on top of a snow covered? \textbf{A}: \textbf{mountain}}
\end{center}
\end{minipage}
\begin{minipage}{0.22\linewidth}
\begin{center}
%\tiny{\textcolor{black}{
%\setlength{\fboxsep}{\fsize}
%\attention{0.062}{what}
%\attention{0.062}{is} 
%\attention{0.061}{the}
%\attention{0.059}{man}
%\attention{0.058}{holding}
%\attention{0.058}{a}
%\attention{0.121}{snowboard}
%\attention{0.055}{on}
%\attention{0.098}{top}
%\attention{0.066}{of}
%\attention{0.058}{a}
%\attention{0.090}{snow}
%\attention{0.098}{covered}
%\attention{0.055}{?}
%}}
\tiny{\textcolor{black}{
\setlength{\fboxsep}{\fsize}
\textcolor[rgb]{0.0000,0.4098,1.0000}{what}
\textcolor[rgb]{0.0000,0.4098,1.0000}{is}}
\textcolor[rgb]{0.0000,0.3941,1.0000}{the}
\textcolor[rgb]{0.0000,0.3627,1.0000}{man}
\textcolor[rgb]{0.0000,0.3471,1.0000}{holding}
\textcolor[rgb]{0.0000,0.3471,1.0000}{a}
\textcolor[rgb]{0.3004,1.0000,0.6673}{snowboard}
\textcolor[rgb]{0.0000,0.3000,1.0000}{on}
\textcolor[rgb]{0.0348,0.9431,0.9330}{top}
\textcolor[rgb]{0.0000,0.4569,1.0000}{of}
\textcolor[rgb]{0.0000,0.3471,1.0000}{a}
\textcolor[rgb]{0.0000,0.8176,1.0000}{snow}
\textcolor[rgb]{0.0348,0.9431,0.9330}{covered}
}
\end{center}
\end{minipage}
\begin{minipage}{0.22\linewidth}
\begin{center}
\tiny{\textcolor{black}{
\setlength{\fboxsep}{\fsize}
\textcolor[rgb]{0.0000,0.3627,1.0000}{what}
\textcolor[rgb]{0.0000,0.3784,1.0000}{is}}
\textcolor[rgb]{0.0000,0.3627,1.0000}{the}
\textcolor[rgb]{0.0000,0.4098,1.0000}{man}
\textcolor[rgb]{0.0000,0.6608,1.0000}{holding}
\textcolor[rgb]{0.0000,0.4725,1.0000}{a}
\textcolor[rgb]{0.0000,0.8333,1.0000}{snowboard}
\textcolor[rgb]{0.0000,0.7078,1.0000}{on}
\textcolor[rgb]{0.0000,0.6294,1.0000}{top}
\textcolor[rgb]{0.0000,0.4412,1.0000}{of}
\textcolor[rgb]{0.0000,0.3784,1.0000}{a}
\textcolor[rgb]{0.0000,0.6608,1.0000}{snow}
\textcolor[rgb]{0.0095,0.9118,0.9583}{covered}
\textcolor[rgb]{0.0000,0.4725,1.0000}{?}
}
\end{center}
\end{minipage}
\begin{minipage}{0.22\linewidth}
\begin{center}
\tiny{\textcolor{black}{
\setlength{\fboxsep}{\fsize}
%\attention{0.076}{what}
%\attention{0.075}{is} 
%\attention{0.069}{the}
%\attention{0.064}{man}
%\attention{0.071}{holding}
%\attention{0.065}{a}
%\attention{0.070}{snowboard}
%\attention{0.071}{on}
%\attention{0.072}{top}
%\attention{0.073}{of}
%\attention{0.074}{a}
%\attention{0.073}{snow}
%\attention{0.070}{covered}
%\attention{0.076}{?}
\textcolor[rgb]{0.0000,0.6294,1.0000}{what}
\textcolor[rgb]{0.0000,0.6137,1.0000}{is}}
\textcolor[rgb]{0.0000,0.5353,1.0000}{the}
\textcolor[rgb]{0.0000,0.4569,1.0000}{man}
\textcolor[rgb]{0.0000,0.5510,1.0000}{holding}
\textcolor[rgb]{0.0000,0.4725,1.0000}{a}
\textcolor[rgb]{0.0000,0.5510,1.0000}{snowboard}
\textcolor[rgb]{0.0000,0.5510,1.0000}{on}
\textcolor[rgb]{0.0000,0.5667,1.0000}{top}
\textcolor[rgb]{0.0000,0.5824,1.0000}{of}
\textcolor[rgb]{0.0000,0.5980,1.0000}{a}
\textcolor[rgb]{0.0000,0.5824,1.0000}{snow}
\textcolor[rgb]{0.0000,0.5510,1.0000}{covered}
\textcolor[rgb]{0.0000,0.6294,1.0000}{?}
}
\end{center}
\end{minipage}
\\
\begin{minipage}{0.22\linewidth}
\includegraphics[width=1\linewidth,  height=0.7\linewidth, scale=0.8, trim={0 0in 0 0in}, clip]{demo/coco_qa/img_230.jpg}
\end{minipage}
\begin{minipage}{0.22\linewidth}
\includegraphics[width=1\linewidth,  height=0.7\linewidth, scale=0.8, trim={0 0in 0 0in}, clip]{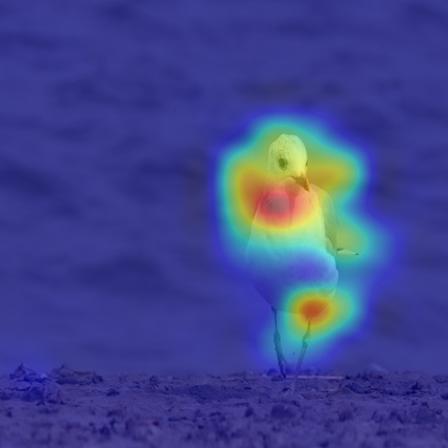}
\end{minipage}
\begin{minipage}{0.22\linewidth}
\includegraphics[width=1\linewidth,  height=0.7\linewidth, scale=0.8, trim={0 0in 0 0in}, clip]{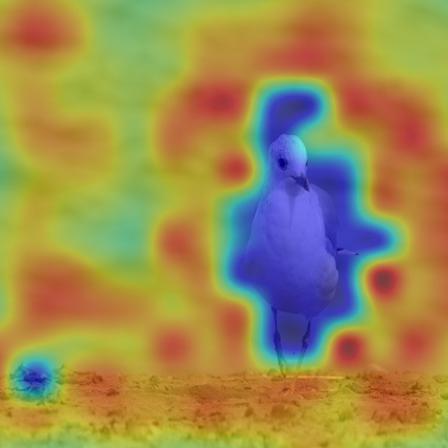}
\end{minipage}
\begin{minipage}{0.22\linewidth}
\includegraphics[width=1\linewidth,  height=0.7\linewidth, scale=0.8, trim={0 0in 0 0in}, clip]{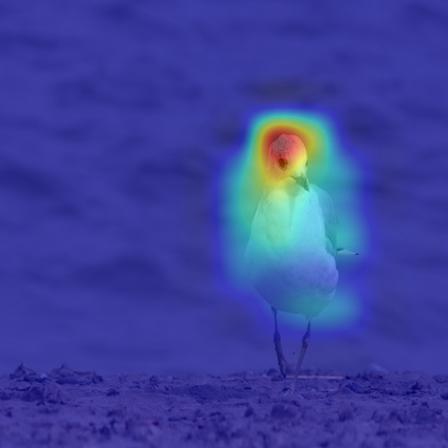}
\end{minipage}
\\
\begin{minipage}{0.22\linewidth}
\begin{center}
\tiny{\textbf{Q}: what is the color of the bird? \textbf{A}: \textbf{white}}
\end{center}
\end{minipage}
\begin{minipage}{0.22\linewidth}
\begin{center}
\tiny{
\setlength{\fboxsep}{\fsize}
%\attention{0.108}{what}
%\attention{0.115}{is}
%\attention{0.117}{the}
%\attention{0.225}{color}
%\attention{0.110}{of}
%\attention{0.117}{the}
%\attention{0.106}{bird}
%\attention{0.113}{?}
\textcolor[rgb]{0.0000,0.6765,1.0000}{what}
\textcolor[rgb]{0.0000,0.7549,1.0000}{is}
\textcolor[rgb]{0.0000,0.7706,1.0000}{the}
\textcolor[rgb]{0.7685,1.0000,0.2292}{color}
\textcolor[rgb]{0.0000,0.6922,1.0000}{of}
\textcolor[rgb]{0.0000,0.7706,1.0000}{the}
\textcolor[rgb]{0.0000,0.6451,1.0000}{bird}
\textcolor[rgb]{0.0000,0.7235,1.0000}{?}
}
\end{center}
\end{minipage}
\begin{minipage}{0.22\linewidth}
\begin{center}
\tiny{
\setlength{\fboxsep}{\fsize}
%\attention{0.061}{what}
%\attention{0.060}{is}
%\attention{0.061}{the}
%\attention{0.576}{color}
%\attention{0.061}{of}
%\attention{0.060}{the}
%\attention{0.061}{bird}
%\attention{0.061}{?}
\textcolor[rgb]{0.0000,0.0000,0.9635}{what}
\textcolor[rgb]{0.0000,0.0000,0.9456}{is}
\textcolor[rgb]{0.0000,0.0000,0.9635}{the}
\textcolor[rgb]{0.6604,0.0000,0.0000}{color}
\textcolor[rgb]{0.0000,0.0000,0.9635}{of}
\textcolor[rgb]{0.0000,0.0000,0.9456}{the}
\textcolor[rgb]{0.0000,0.0000,0.9635}{bird}
\textcolor[rgb]{0.0000,0.0000,0.9635}{?}
}
\end{center}
\end{minipage}
\begin{minipage}{0.22\linewidth}
\begin{center}
\tiny{
\setlength{\fboxsep}{\fsize}
%\attention{0.020}{what}
%\attention{0.020}{is}
%\attention{0.021}{the}
%\attention{0.222}{color}
%\attention{0.222}{of}
%\attention{0.181}{the}
%\attention{0.185}{bird}
%\attention{0.220}{?}
\textcolor[rgb]{0.0000,0.0000,0.7139}{what}
\textcolor[rgb]{0.0000,0.0000,0.7139}{is}
\textcolor[rgb]{0.0000,0.0000,0.7139}{the}
\textcolor[rgb]{0.3384,1.0000,0.6293}{color}
\textcolor[rgb]{0.3384,1.0000,0.6293}{of}
\textcolor[rgb]{0.2498,1.0000,0.7179}{the}
\textcolor[rgb]{0.2878,1.0000,0.6799}{bird}
\textcolor[rgb]{0.3257,1.0000,0.6420}{?}
}
\end{center}
\end{minipage}
\\
\begin{minipage}{0.22\linewidth}
\includegraphics[width=1\linewidth,  height=0.7\linewidth, scale=0.8, trim={0 0in 0 0in}, clip]{demo/coco_qa/img_133.jpg}
\end{minipage}
\begin{minipage}{0.22\linewidth}
\includegraphics[width=1\linewidth,  height=0.7\linewidth, scale=0.8, trim={0 0in 0 0in}, clip]{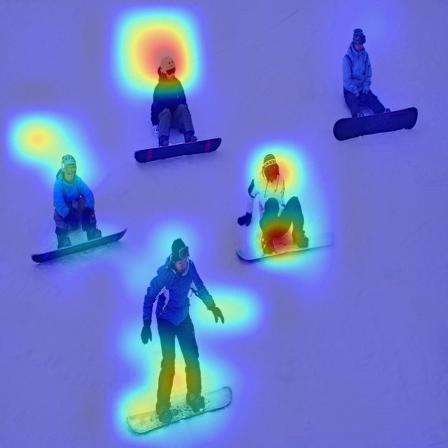}
\end{minipage}
\begin{minipage}{0.22\linewidth}
\includegraphics[width=1\linewidth,  height=0.7\linewidth, scale=0.8, trim={0 0in 0 0in}, clip]{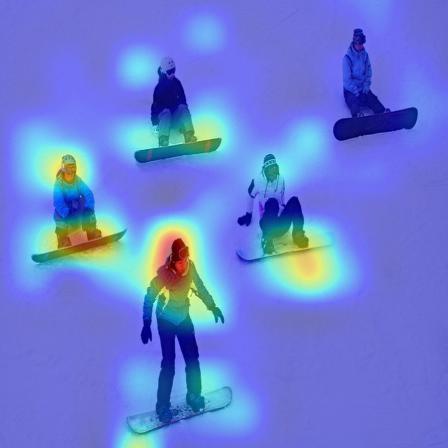}
\end{minipage}
\begin{minipage}{0.22\linewidth}
\includegraphics[width=1\linewidth,  height=0.7\linewidth, scale=0.8, trim={0 0in 0 0in}, clip]{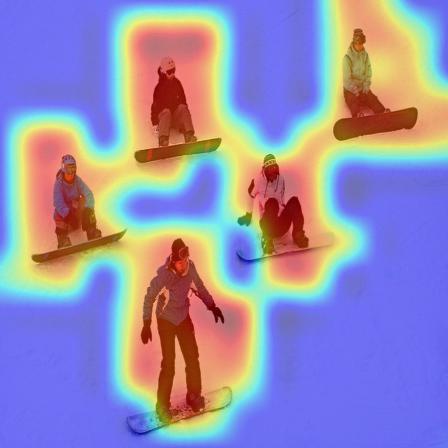}
\end{minipage}
\\
\begin{minipage}{0.22\linewidth}
\begin{center}
\tiny{\textbf{Q}: how many snowboarders in formation in the snow, four is sitting? \textbf{A}: \textbf{5}}
\end{center}
\end{minipage}
\begin{minipage}{0.22\linewidth}
\begin{center}
\tiny{
\setlength{\fboxsep}{\fsize}
%\attention{0.114}{how}
%\attention{0.078}{many}
%\attention{0.134}{snowboarders}
%\attention{0.048}{in}
%\attention{0.070}{formation}
%\attention{0.048}{in}
%\attention{0.066}{the}
%\attention{0.110}{snow}
%\attention{0.051}{,}
%\attention{0.055}{four}
%\attention{0.051}{is}
%\attention{0.123}{sitting}
%\attention{0.053}{?}
\textcolor[rgb]{0.0980,1.0000,0.8697}{how}
\textcolor[rgb]{0.0000,0.5353,1.0000}{many}
\textcolor[rgb]{0.3131,1.0000,0.6546}{snowboarders}
\textcolor[rgb]{0.0000,0.1431,1.0000}{in}
\textcolor[rgb]{0.0000,0.4412,1.0000}{formation}
\textcolor[rgb]{0.0000,0.1431,1.0000}{in}
\textcolor[rgb]{0.0000,0.3784,1.0000}{the}
\textcolor[rgb]{0.0601,0.9745,0.9077}{snow}
\textcolor[rgb]{0.0000,0.1745,1.0000}{,}
\textcolor[rgb]{0.0000,0.2373,1.0000}{four}
\textcolor[rgb]{0.0000,0.1745,1.0000}{is}
\textcolor[rgb]{0.2292,1.0000,0.7685}{sitting}
\textcolor[rgb]{0.0000,0.2259,1.0000}{?}
}
\end{center}
\end{minipage}
\begin{minipage}{0.22\linewidth}
\begin{center}
\tiny{
\setlength{\fboxsep}{\fsize}
%\attention{0.130}{how}
%\attention{0.266}{many}
%\attention{0.107}{snowboarders}
%\attention{0.043}{in}
%\attention{0.037}{formation}
%\attention{0.038}{in}
%\attention{0.037}{the}
%\attention{0.091}{snow}
%\attention{0.070}{,}
%\attention{0.059}{four}
%\attention{0.038}{is}
%\attention{0.048}{sitting}
%\attention{0.036}{?}
\textcolor[rgb]{0.0474,0.9588,0.9203}{how}
\textcolor[rgb]{1.0000,0.5962,0.0000}{many}
\textcolor[rgb]{0.0000,0.6922,1.0000}{snowboarders}
\textcolor[rgb]{0.0000,0.0000,1.0000}{in}
\textcolor[rgb]{0.0000,0.0000,0.9635}{formation}
\textcolor[rgb]{0.0000,0.0000,0.9813}{in}
\textcolor[rgb]{0.0000,0.0000,0.9635}{the}
\textcolor[rgb]{0.0000,0.5196,1.0000}{snow}
\textcolor[rgb]{0.0000,0.2843,1.0000}{,}
\textcolor[rgb]{0.0000,0.1588,1.0000}{four}
\textcolor[rgb]{0.0000,0.0000,0.9813}{is}
\textcolor[rgb]{0.0000,0.0333,1.0000}{sitting}
\textcolor[rgb]{0.0000,0.0000,0.9456}{?}
}
\end{center}
\end{minipage}
\begin{minipage}{0.22\linewidth}
\begin{center}
\tiny{
\setlength{\fboxsep}{\fsize}
%\attention{0.079}{how}
%\attention{0.078}{many}
%\attention{0.078}{snowboarders}
%\attention{0.078}{in}
%\attention{0.074}{formation}
%\attention{0.074}{in}
%\attention{0.077}{the}
%\attention{0.078}{snow}
%\attention{0.078}{,}
%\attention{0.078}{four}
%\attention{0.074}{is}
%\attention{0.078}{sitting}
%\attention{0.077}{?}
\textcolor[rgb]{0.0000,0.6294,1.0000}{how}
\textcolor[rgb]{0.0000,0.6137,1.0000}{many}
\textcolor[rgb]{0.0000,0.6137,1.0000}{snowboarders}
\textcolor[rgb]{0.0000,0.6137,1.0000}{in}
\textcolor[rgb]{0.0000,0.5667,1.0000}{formation}
\textcolor[rgb]{0.0000,0.5667,1.0000}{in}
\textcolor[rgb]{0.0000,0.5980,1.0000}{the}
\textcolor[rgb]{0.0000,0.6137,1.0000}{snow}
\textcolor[rgb]{0.0000,0.6137,1.0000}{,}
\textcolor[rgb]{0.0000,0.6137,1.0000}{four}
\textcolor[rgb]{0.0000,0.5667,1.0000}{is}
\textcolor[rgb]{0.0000,0.6137,1.0000}{sitting}
\textcolor[rgb]{0.0000,0.5980,1.0000}{?}
}
\end{center}
\end{minipage}
\caption{Visualization of image and question co-attention maps on the COCO-QA dataset. From left to right: original image and question pairs, word level co-attention maps, phrase level co-attention maps and question level co-attention maps. For visualization, both image and question attentions are scaled (from red:high to blue:low). Best viewed in color.}
\label{fig:vis}
\end{figure}

We now visualize some co-attention maps generated by our method in Fig.~\ref{fig:vis}. At the word level, our model attends mostly to the object regions in an image, e.g., heads, bird. At the phrase level, the image attention has different patterns across images. For the first two images, the attention transfers from objects to background regions. For the third image, the attention becomes more focused on the objects. We suspect that this is caused by the different question types. On the question side, our model is capable of localizing the key phrases in the question, thus essentially discovering the question types in the dataset. For example, our model pays attention to the phrases ``what color'' and ``how many snowboarders''. Our model successfully attends to the regions in images and phrases in the questions appropriate for answering the question, e.g., ``color of the bird'' and bird region.  Because our model performs co-attention at three levels, it often captures complementary information from each level, and then combines them to predict the answer.
%%%%%%%%%%%%%%%%%%%%%%%%%%%%%%%%%%%%%%%%%%%%%%%%%%%%%%%%%%%
\section{Conclusion}
\label{sec:con}
\vspace*{-2mm}
%%%%%%%%%%%%%%%%%%%%%%%%%%%%%%%%%%%%%%%%%%%%%%%%%%%%%%%%%%%
In this paper, we proposed a hierarchical co-attention model for visual question answering. Co-attention allows our model to attend to different regions of the image as well as different fragments of the question. We model the question hierarchically at three levels to capture information from different granularities. The ablation studies further demonstrate the roles of co-attention and question hierarchy in our final performance. Through visualizations, we can see that our model co-attends to interpretable regions of images and questions for predicting the answer. Though our model was evaluated on visual question answering, it can be potentially applied to other tasks involving vision and language.

\small{\textbf{Acknowledgements}}

\small{This work was funded in part by NSF CAREER awards to DP and DB, an ONR YIP award to DP, ONR Grant N00014-14-1-0679 to DB, a Sloan Fellowship to DP, ARO YIP awards to DB and DP, a Allen Distinguished Investigator award to DP from the Paul G. Allen Family Foundation, ICTAS Junior Faculty awards to DB and DP, Google Faculty Research Awards to DP and DB, AWS in Education Research grant to DB, and NVIDIA GPU donations to DB. The views and conclusions contained herein are those of the authors and should not be interpreted as necessarily representing the official policies or endorsements, either expressed or implied, of the U.S. Government or any sponsor. 
}
%%%%%%%%%%%%%%%%%%%%%%%%%%%%%%%%%%%%%%%%%%%%%%%%%%%%%%%%%%%
%%%%%%%%%%%%%%%%%%%%%%%%%%%%%%%%%%%%%%%%%%%%%%%%%%%%%%%%%%%
\begin{figure}[p]
\begin{minipage}{0.19\linewidth}
\includegraphics[width=1\linewidth,  height=0.7\linewidth, scale=0.8, trim={0 0in 0 0in}, clip]{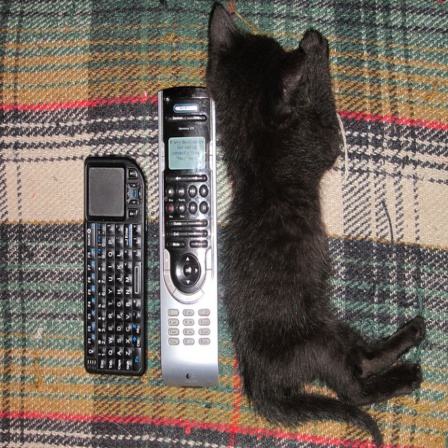}
\end{minipage}
\begin{minipage}{0.19\linewidth}
\includegraphics[width=1\linewidth,  height=0.7\linewidth, scale=0.8, trim={0 0in 0 0in}, clip]{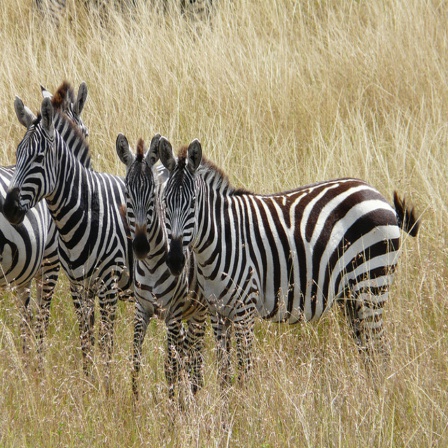}
\end{minipage}
\begin{minipage}{0.19\linewidth}
\includegraphics[width=1\linewidth,  height=0.7\linewidth, scale=0.8, trim={0 0in 0 0in}, clip]{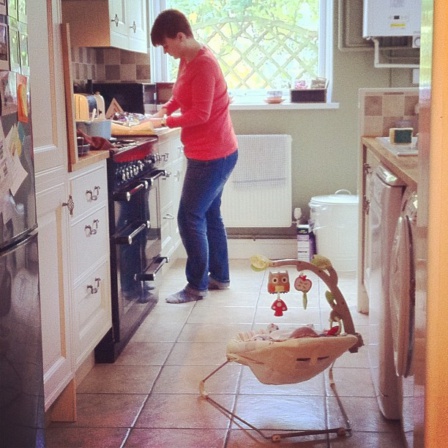}
\end{minipage}
\begin{minipage}{0.19\linewidth}
\includegraphics[width=1\linewidth,  height=0.7\linewidth, scale=0.8, trim={0 0in 0 0in}, clip]{demo/vqa/img_116.jpg}
\end{minipage}
\begin{minipage}{0.19\linewidth}
\includegraphics[width=1\linewidth,  height=0.7\linewidth, scale=0.8, trim={0 0in 0 0in}, clip]{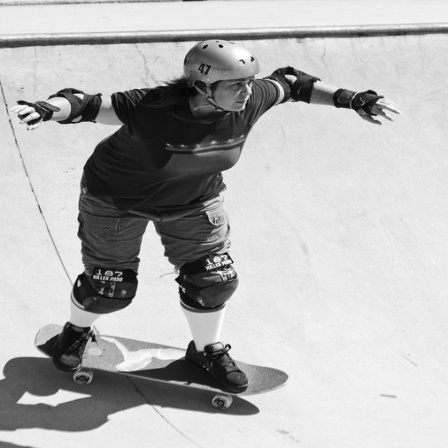}
\end{minipage}

\begin{minipage}{0.19\linewidth}
\begin{center}
\tiny{\textbf{Q}: what is the color of the kitten? \textbf{A}: \textbf{black}}
\end{center}
\end{minipage}
\begin{minipage}{0.19\linewidth}
\begin{center}
\tiny{\textbf{Q}: what are standing in tall dry grass look at the tourists? \textbf{A}: \textbf{zebras}}
\end{center}
\end{minipage}
\begin{minipage}{0.19\linewidth}
\begin{center}
\tiny{\textbf{Q}: where is the woman while her baby is sleeping? \textbf{A}: \textbf{kitchen}}
\end{center}
\end{minipage}
\begin{minipage}{0.19\linewidth}
\begin{center}
\tiny{\textbf{Q}:  what seating area is on the right? \textbf{A}: \textbf{park}}
\end{center}
\end{minipage}
\begin{minipage}{0.19\linewidth}
\begin{center}
\tiny{\textbf{Q}: is the person dressed properly for this sport? \textbf{A}: \textbf{yes}}
\end{center}   
\end{minipage}

\begin{minipage}{0.19\linewidth}
\includegraphics[width=1\linewidth,  height=0.7\linewidth, scale=0.8, trim={0 0in 0 0in}, clip]{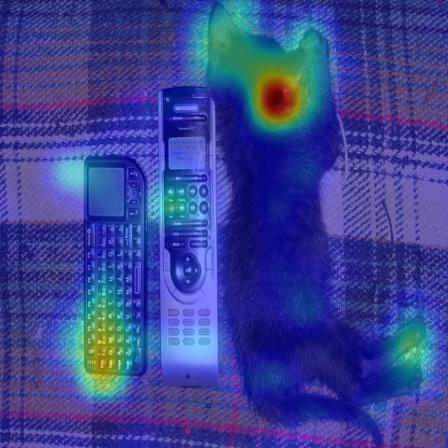}
\end{minipage}
\begin{minipage}{0.19\linewidth}
\includegraphics[width=1\linewidth,  height=0.7\linewidth, scale=0.8, trim={0 0in 0 0in}, clip]{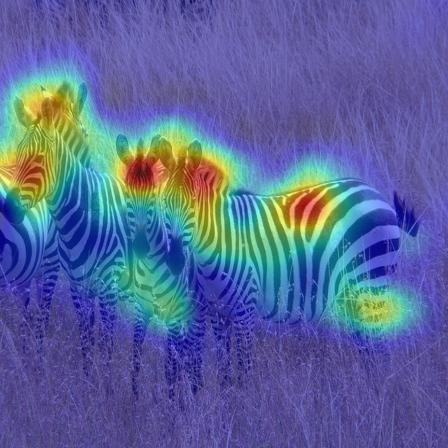}
\end{minipage}
\begin{minipage}{0.19\linewidth}
\includegraphics[width=1\linewidth,  height=0.7\linewidth, scale=0.8, trim={0 0in 0 0in}, clip]{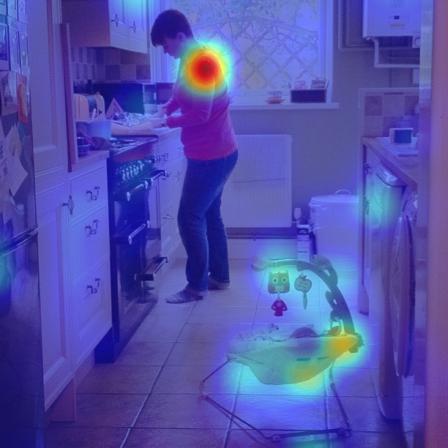}
\end{minipage}
\begin{minipage}{0.19\linewidth}
\includegraphics[width=1\linewidth,  height=0.7\linewidth, scale=0.8, trim={0 0in 0 0in}, clip]{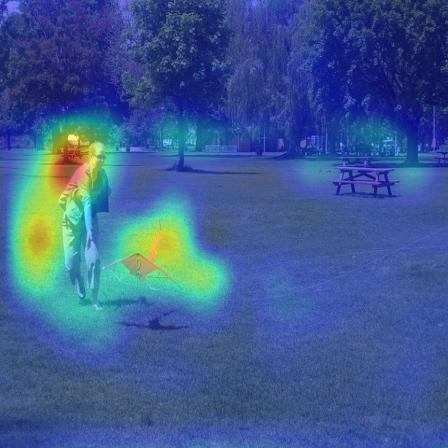}
\end{minipage}
\begin{minipage}{0.19\linewidth}
\includegraphics[width=1\linewidth,  height=0.7\linewidth, scale=0.8, trim={0 0in 0 0in}, clip]{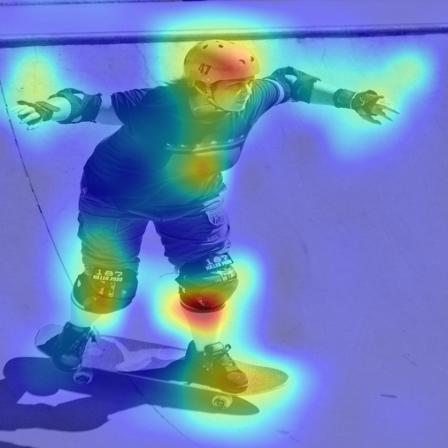}
\end{minipage}

\begin{minipage}{0.19\linewidth}
\begin{center}
%\attention{0.125}{what}
%\attention{0.088}{is} 
%\attention{0.150}{the}
%\attention{0.139}{color}
%\attention{0.083}{of}
%\attention{0.150}{the}
%\attention{0.116 }{kitten}
%\attention{0.149}{?}
%}}
\tiny{\textcolor{black}{
\setlength{\fboxsep}{\fsize}
\textcolor[rgb]{0.0000,0.8804,0.9836}{what}
\textcolor[rgb]{0.0000,0.4725,1.0000}{is}}
\textcolor[rgb]{0.2119,1.0000,0.7559}{the}
\textcolor[rgb]{0.1107,1.0000,0.8571}{color}
\textcolor[rgb]{0.0000,0.4098,1.0000}{of}
\textcolor[rgb]{0.2119,1.0000,0.7559}{the}
\textcolor[rgb]{0.0000,0.7863,1.0000}{kitten}
\textcolor[rgb]{0.1992,1.0000,0.7685}{?}
}
\end{center}
\end{minipage}
\begin{minipage}{0.19\linewidth}
\begin{center}
\tiny{
\setlength{\fboxsep}{\fsize}
%\attention{0.039}{what}
%\attention{0.060}{are}
%\attention{0.128}{standing}
%\attention{0.091}{in}
%\attention{0.132}{tall}
%\attention{0.110}{dry}
%\attention{0.127}{grass}
%\attention{0.080}{look}	
%\attention{0.060}{at}
%\attention{0.060}{the}
%\attention{0.050}{tourists}
%\attention{0.062}{?}
\textcolor[rgb]{0.0000,0.0020,1.0000}{what}
\textcolor[rgb]{0.0000,0.2686,1.0000}{are}
\textcolor[rgb]{0.2119,1.0000,0.7559}{standing}
\textcolor[rgb]{0.0000,0.6765,1.0000}{in}
\textcolor[rgb]{0.2498,1.0000,0.7179}{tall}
\textcolor[rgb]{0.0221,0.9275,0.9456}{dry}
\textcolor[rgb]{0.1992,1.0000,0.7685}{grass}
\textcolor[rgb]{0.0000,0.5353,1.0000}{look}  
\textcolor[rgb]{0.0000,0.2686,1.0000}{at}
\textcolor[rgb]{0.0000,0.2686,1.0000}{the}
\textcolor[rgb]{0.0000,0.1431,1.0000}{tourists}
\textcolor[rgb]{0.0000,0.3000,1.0000}{?}
}
\end{center}
\end{minipage}
\begin{minipage}{0.19\linewidth}
\begin{center}
\tiny{
\setlength{\fboxsep}{\fsize}
%\attention{0.072}{where}
%\attention{0.078}{is}
%\attention{0.076}{the}
%\attention{0.090}{woman}
%\attention{0.077}{while}
%\attention{0.104}{her}
%\attention{0.139}{baby}
%\attention{0.078}{is}
%\attention{0.213}{sleeping}
%\attention{0.072}{?}
\textcolor[rgb]{0.0000,0.3314,1.0000}{where}
\textcolor[rgb]{0.0000,0.4098,1.0000}{is}
\textcolor[rgb]{0.0000,0.3784,1.0000}{the}
\textcolor[rgb]{0.0000,0.5510,1.0000}{woman}
\textcolor[rgb]{0.0000,0.3941,1.0000}{while}
\textcolor[rgb]{0.0000,0.7078,1.0000}{her}
\textcolor[rgb]{0.1739,1.0000,0.7938}{baby}
\textcolor[rgb]{0.0000,0.4098,1.0000}{is}
\textcolor[rgb]{0.8697,1.0000,0.0980}{sleeping}
\textcolor[rgb]{0.0000,0.3314,1.0000}{?}
}
\end{center}
\end{minipage}
\begin{minipage}{0.19\linewidth}
\begin{center}
\tiny{
\setlength{\fboxsep}{\fsize}
%\attention{0.086}{what}
%\attention{0.413}{seating}
%\attention{0.241}{area}
%\attention{0.036}{is}
%\attention{0.049}{on}
%\attention{0.039}{the}
%\attention{0.104}{right}
%\attention{0.031}{?}
\textcolor[rgb]{0.0000,0.1745,1.0000}{what}
\textcolor[rgb]{1.0000,0.3203,0.0000}{seating}
\textcolor[rgb]{0.4143,1.0000,0.5534}{area}
\textcolor[rgb]{0.0000,0.0000,0.8209}{is}
\textcolor[rgb]{0.0000,0.0000,0.9278}{on}
\textcolor[rgb]{0.0000,0.0000,0.8387}{the}
\textcolor[rgb]{0.0000,0.3157,1.0000}{right}
\textcolor[rgb]{0.0000,0.0000,0.7674}{?}
}
\end{center}
\end{minipage}
\begin{minipage}{0.19\linewidth}
\begin{center}
\tiny{
\setlength{\fboxsep}{\fsize}
%\attention{0.021}{is}
%\attention{0.018}{the}
%\attention{0.019}{person}
%\attention{0.218}{dressed}
%\attention{0.093}{properly}
%\attention{0.409}{for}
%\attention{0.036}{the}
%\attention{0.167}{sport}
%\attention{0.019}{?}
\textcolor[rgb]{0.0000,0.0000,0.6783}{is}
\textcolor[rgb]{0.0000,0.0000,0.6604}{the}
\textcolor[rgb]{0.0000,0.0000,0.6604}{person}
\textcolor[rgb]{0.2625,1.0000,0.7052}{dressed}
\textcolor[rgb]{0.0000,0.2373,1.0000}{properly}
\textcolor[rgb]{1.0000,0.3638,0.0000}{for}
\textcolor[rgb]{0.0000,0.0000,0.8209}{the}
\textcolor[rgb]{0.0000,0.8176,1.0000}{sport}
\textcolor[rgb]{0.0000,0.0000,0.6604}{?}
}
\end{center}   
\end{minipage}

\begin{minipage}{0.19\linewidth}
\includegraphics[width=1\linewidth,  height=0.7\linewidth, scale=0.8, trim={0 0in 0 0in}, clip]{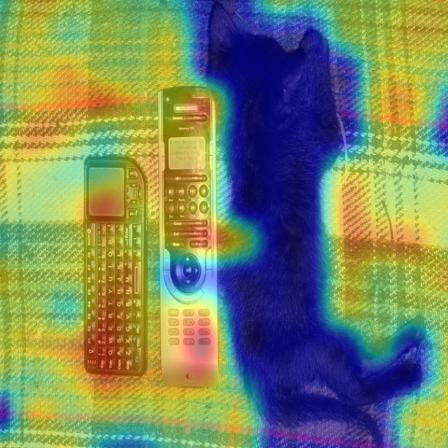}
\end{minipage}
\begin{minipage}{0.19\linewidth}
\includegraphics[width=1\linewidth,  height=0.7\linewidth, scale=0.8, trim={0 0in 0 0in}, clip]{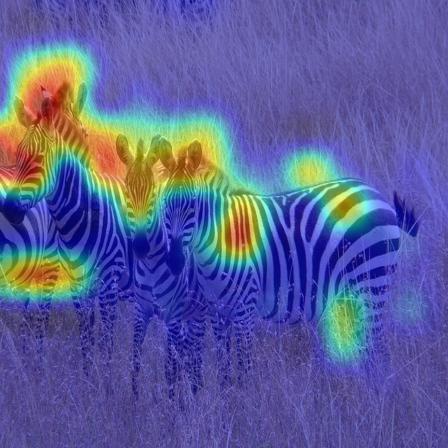}
\end{minipage}
\begin{minipage}{0.19\linewidth}
\includegraphics[width=1\linewidth,  height=0.7\linewidth, scale=0.8, trim={0 0in 0 0in}, clip]{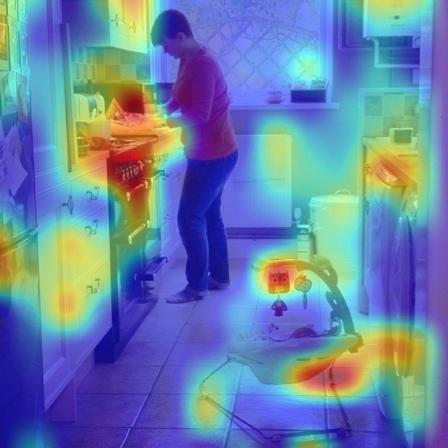}
\end{minipage}
\begin{minipage}{0.19\linewidth}
\includegraphics[width=1\linewidth,  height=0.7\linewidth, scale=0.8, trim={0 0in 0 0in}, clip]{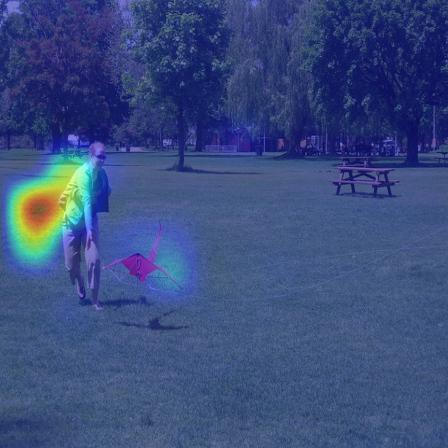}
\end{minipage}
\begin{minipage}{0.19\linewidth}
\includegraphics[width=1\linewidth,  height=0.7\linewidth, scale=0.8, trim={0 0in 0 0in}, clip]{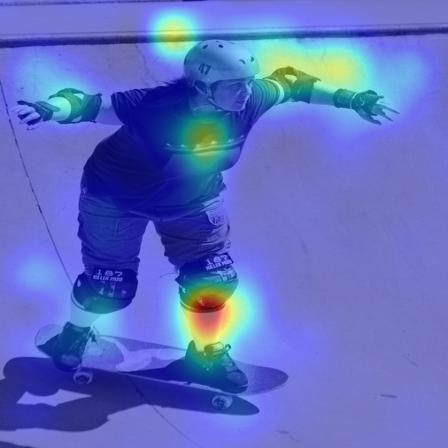}
\end{minipage}

\vspace{0pt}
\begin{minipage}{0.19\linewidth}
\begin{center}
\tiny{\textcolor{black}{
\setlength{\fboxsep}{\fsize}
%\attention{0.047}{what}
%\attention{0.049}{is} 
%\attention{0.043}{the}
%\attention{0.666}{color}
%\attention{0.051}{of}
%\attention{0.043}{the}
%\attention{0.050 }{kitten}
%\attention{0.051}{?}
%}}
\textcolor[rgb]{0.0000,0.0000,0.8030}{what}
\textcolor[rgb]{0.0000,0.0000,0.8209}{is}}
\textcolor[rgb]{0.0000,0.0000,0.7852}{the}
\textcolor[rgb]{0.5713,0.0000,0.0000}{color}
\textcolor[rgb]{0.0000,0.0000,0.8387}{of}
\textcolor[rgb]{0.0000,0.0000,0.7852}{the}
\textcolor[rgb]{0.0000,0.0000,0.8209}{kitten}
\textcolor[rgb]{0.0000,0.0000,0.8387}{?}
}
\end{center}
\end{minipage}
\begin{minipage}{0.19\linewidth}
\begin{center}
\tiny{
\setlength{\fboxsep}{\fsize}
%\attention{0.061}{what}
%\attention{0.062}{are}
%\attention{0.074}{standing}
%\attention{0.075}{in}
%\attention{0.075}{tall}
%\attention{0.081}{dry}
%\attention{0.080}{grass}
%\attention{0.208}{look}	
%\attention{0.073}{at}
%\attention{0.073}{the}
%\attention{0.066}{tourists}
%\attention{0.073}{?}
\textcolor[rgb]{0.0000,0.2686,1.0000}{what}
\textcolor[rgb]{0.0000,0.2686,1.0000}{are}
\textcolor[rgb]{0.0000,0.4255,1.0000}{standing}
\textcolor[rgb]{0.0000,0.4412,1.0000}{in}
\textcolor[rgb]{0.0000,0.4412,1.0000}{tall}
\textcolor[rgb]{0.0000,0.5196,1.0000}{dry}
\textcolor[rgb]{0.0000,0.5039,1.0000}{grass}
\textcolor[rgb]{0.9836,0.9448,0.0000}{look}
\textcolor[rgb]{0.0000,0.4098,1.0000}{at}
\textcolor[rgb]{0.0000,0.4098,1.0000}{the}
\textcolor[rgb]{0.0000,0.3314,1.0000}{tourists}
\textcolor[rgb]{0.0000,0.4098,1.0000}{?}
}
\end{center}
\end{minipage}
\begin{minipage}{0.19\linewidth}
\begin{center}
\tiny{
\setlength{\fboxsep}{\fsize}
%\attention{0.057}{where}
%\attention{0.055}{is}
%\attention{0.053}{the}
%\attention{0.071}{woman}
%\attention{0.183}{while}
%\attention{0.119}{her}
%\attention{0.193}{baby}
%\attention{0.056}{is}
%\attention{0.132}{sleeping}
%\attention{0.081}{?}
\textcolor[rgb]{0.0000,0.1431,1.0000}{where}
\textcolor[rgb]{0.0000,0.1118,1.0000}{is}
\textcolor[rgb]{0.0000,0.0961,1.0000}{the}
\textcolor[rgb]{0.0000,0.3000,1.0000}{woman}
\textcolor[rgb]{0.5281,1.0000,0.4396}{while}
\textcolor[rgb]{0.0000,0.8333,1.0000}{her}
\textcolor[rgb]{0.6293,1.0000,0.3384}{baby}
\textcolor[rgb]{0.0000,0.1275,1.0000}{is}
\textcolor[rgb]{0.0727,0.9902,0.8950}{sleeping}
\textcolor[rgb]{0.0000,0.4098,1.0000}{?}
}
\end{center}
\end{minipage}
\begin{minipage}{0.19\linewidth}
\begin{center}
\tiny{
\setlength{\fboxsep}{\fsize}
%\attention{0.075}{what}
%\attention{0.295}{seating}
%\attention{0.356}{area}
%\attention{0.047}{is}
%\attention{0.031}{on}
%\attention{0.027}{the}
%\attention{0.110}{right}
%\attention{0.058}{?}
\textcolor[rgb]{0.0000,0.1118,1.0000}{what}
\textcolor[rgb]{0.8191,1.0000,0.1486}{seating}
\textcolor[rgb]{1.0000,0.6688,0.0000}{area}
\textcolor[rgb]{0.0000,0.0000,0.9278}{is}
\textcolor[rgb]{0.0000,0.0000,0.7852}{on}
\textcolor[rgb]{0.0000,0.0000,0.7496}{the}
\textcolor[rgb]{0.0000,0.3941,1.0000}{right}
\textcolor[rgb]{0.0000,0.0000,1.0000}{?}
}
\end{center}
\end{minipage}
\begin{minipage}{0.19\linewidth}
\begin{center}
\tiny{
\setlength{\fboxsep}{\fsize}
%\attention{0.021}{is}
%\attention{0.016}{the}
%\attention{0.019}{person}
%\attention{0.166}{dressed}
%\attention{0.245}{properly}
%\attention{0.192}{for}
%\attention{0.080}{the}
%\attention{0.224}{sport}
%\attention{0.035}{?}
\textcolor[rgb]{0.0000,0.0000,0.7139}{is}
\textcolor[rgb]{0.0000,0.0000,0.6604}{the}
\textcolor[rgb]{0.0000,0.0000,0.6961}{person}
\textcolor[rgb]{0.1233,1.0000,0.8444}{dressed}
\textcolor[rgb]{0.7179,1.0000,0.2498}{properly}
\textcolor[rgb]{0.3131,1.0000,0.6546}{for}
\textcolor[rgb]{0.0000,0.2373,1.0000}{the}
\textcolor[rgb]{0.5534,1.0000,0.4143}{sport}
\textcolor[rgb]{0.0000,0.0000,0.8565}{?}
}
\end{center}   
\end{minipage}

\begin{minipage}{0.19\linewidth}
\includegraphics[width=1\linewidth,  height=0.7\linewidth, scale=0.8, trim={0 0in 0 0in}, clip]{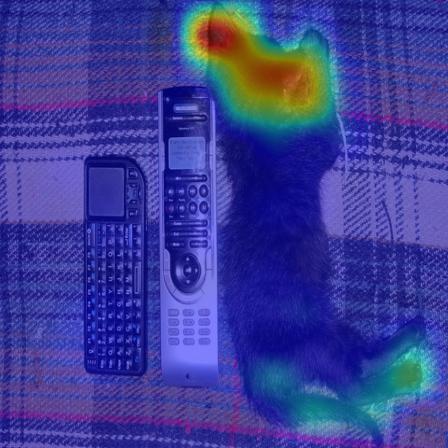}
\end{minipage}
\begin{minipage}{0.19\linewidth}
\includegraphics[width=1\linewidth,  height=0.7\linewidth, scale=0.8, trim={0 0in 0 0in}, clip]{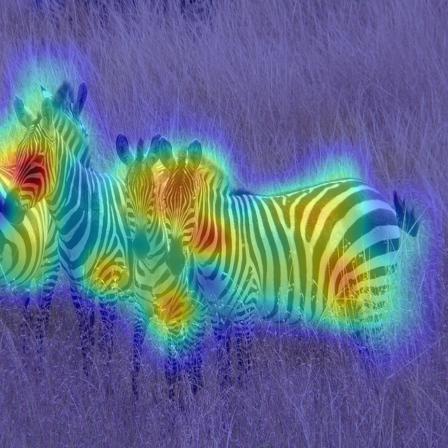}
\end{minipage}
\begin{minipage}{0.19\linewidth}
\includegraphics[width=1\linewidth,  height=0.7\linewidth, scale=0.8, trim={0 0in 0 0in}, clip]{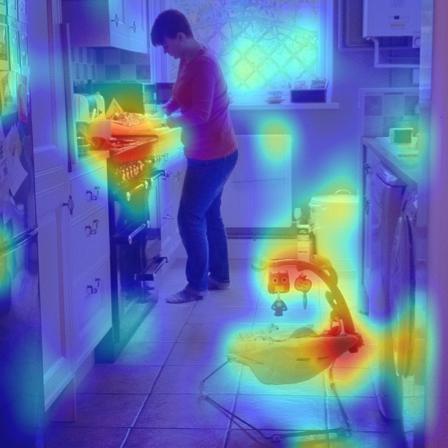}
\end{minipage}
\begin{minipage}{0.19\linewidth}
\includegraphics[width=1\linewidth,  height=0.7\linewidth, scale=0.8, trim={0 0in 0 0in}, clip]{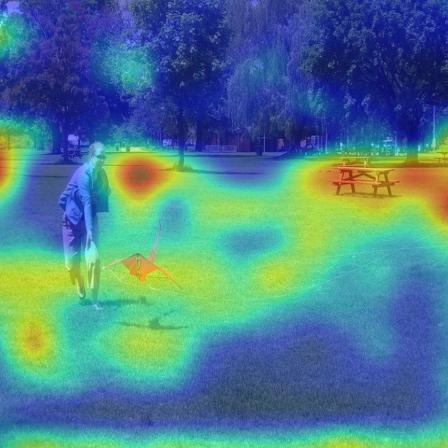}
\end{minipage}
\begin{minipage}{0.19\linewidth}
\includegraphics[width=1\linewidth,  height=0.7\linewidth, scale=0.8, trim={0 0in 0 0in}, clip]{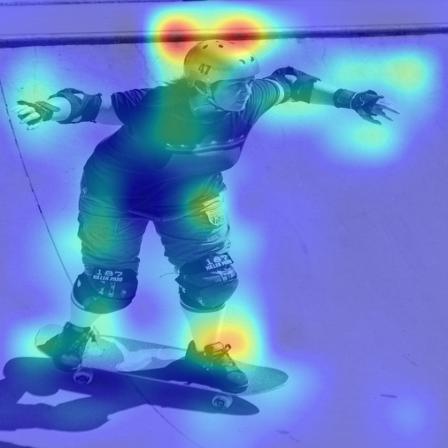}
\end{minipage}

\vspace{0pt}
\begin{minipage}{0.19\linewidth}
\begin{center}
\tiny{\textcolor{black}{
\setlength{\fboxsep}{\fsize}
%\tiny{\textcolor{black}{
%\setlength{\fboxsep}{\fsize}
%\attention{0.021}{what}
%\attention{0.022}{is} 
%\attention{0.021}{the}
%\attention{0.181}{color}
%\attention{0.176}{of}
%\attention{0.189}{the}
%\attention{0.201 }{kitten}
%\attention{0.189}{?}
%}}
\textcolor[rgb]{0.0000,0.0000,0.7139}{what}
\textcolor[rgb]{0.0000,0.0000,0.7317}{is}}
\textcolor[rgb]{0.0000,0.0000,0.7139}{the}
\textcolor[rgb]{0.2625,1.0000,0.7052}{color}
\textcolor[rgb]{0.2245,1.0000,0.7432}{of}
\textcolor[rgb]{0.3257,1.0000,0.6420}{the}
\textcolor[rgb]{0.4143,1.0000,0.5534}{kitten}
\textcolor[rgb]{0.3257,1.0000,0.6420}{?}
}
\end{center}
\end{minipage}
\begin{minipage}{0.19\linewidth}
\begin{center}
\tiny{
\setlength{\fboxsep}{\fsize}
%\attention{0.025}{what}
%\attention{0.025}{are}
%\attention{0.072}{standing}
%\attention{0.092}{in}
%\attention{0.100}{tall}
%\attention{0.101}{dry}
%\attention{0.101}{grass}
%\attention{0.096}{look}    
%\attention{0.101}{at}
%\attention{0.095}{the}
%\attention{0.097}{tourists}
%\attention{0.076}{?}
\textcolor[rgb]{0.0000,0.0000,0.8743}{what}
\textcolor[rgb]{0.0000,0.0000,0.8743}{are}
\textcolor[rgb]{0.0000,0.4569,1.0000}{standing}
\textcolor[rgb]{0.0000,0.7235,1.0000}{in}
\textcolor[rgb]{0.0000,0.8333,1.0000}{tall}
\textcolor[rgb]{0.0000,0.8490,1.0000}{dry}
\textcolor[rgb]{0.0000,0.8490,1.0000}{grass}
\textcolor[rgb]{0.0000,0.7863,1.0000}{look}
\textcolor[rgb]{0.0000,0.8490,1.0000}{at}
\textcolor[rgb]{0.0000,0.7706,1.0000}{the}
\textcolor[rgb]{0.0000,0.8020,1.0000}{tourists}
\textcolor[rgb]{0.0000,0.5196,1.0000}{?}
}
\end{center}
\end{minipage}
\begin{minipage}{0.19\linewidth}
\begin{center}
\tiny{
\setlength{\fboxsep}{\fsize}
%\attention{0.100}{where}
%\attention{0.087}{is}
%\attention{0.088}{the}
%\attention{0.088}{woman}
%\attention{0.100}{while}
%\attention{0.107}{her}
%\attention{0.112}{baby}
%\attention{0.110}{is}
%\attention{0.104}{sleeping}
%\attention{0.104}{?}
\textcolor[rgb]{0.0000,0.7549,1.0000}{where}
\textcolor[rgb]{0.0000,0.5980,1.0000}{is}
\textcolor[rgb]{0.0000,0.5980,1.0000}{the}
\textcolor[rgb]{0.0000,0.5980,1.0000}{woman}
\textcolor[rgb]{0.0000,0.7549,1.0000}{while}
\textcolor[rgb]{0.0000,0.8490,1.0000}{her}
\textcolor[rgb]{0.0095,0.9118,0.9583}{baby}
\textcolor[rgb]{0.0000,0.8804,0.9836}{is}
\textcolor[rgb]{0.0000,0.8020,1.0000}{sleeping}
\textcolor[rgb]{0.0000,0.8020,1.0000}{?}
}
\end{center}
\end{minipage}
\begin{minipage}{0.19\linewidth}
\begin{center}
\tiny{
\setlength{\fboxsep}{\fsize}
%\attention{0.033}{what}
%\attention{0.032}{seating}
%\attention{0.060}{area}
%\attention{0.090}{is}
%\attention{0.148}{on}
%\attention{0.201}{the}
%\attention{0.221}{right}
%\attention{0.215}{?}
\textcolor[rgb]{0.0000,0.0000,0.8565}{what}
\textcolor[rgb]{0.0000,0.0000,0.8387}{seating}
\textcolor[rgb]{0.0000,0.0804,1.0000}{area}
\textcolor[rgb]{0.0000,0.3627,1.0000}{is}
\textcolor[rgb]{0.0221,0.9275,0.9456}{on}
\textcolor[rgb]{0.4396,1.0000,0.5281}{the}
\textcolor[rgb]{0.5914,1.0000,0.3763}{right}
\textcolor[rgb]{0.5534,1.0000,0.4143}{?}
}	
\end{center}
\end{minipage}
\begin{minipage}{0.19\linewidth}
\begin{center}
\tiny{
\setlength{\fboxsep}{\fsize}
%\attention{0.010}{is}
%\attention{0.008}{the}
%\attention{0.008}{person}
%\attention{0.010}{dressed}
%\attention{0.038}{properly}
%\attention{0.071}{for}
%\attention{0.084}{the}
%\attention{0.228}{sport}
%\attention{0.543}{?}
\textcolor[rgb]{0.0000,0.0000,0.5713}{is}
\textcolor[rgb]{0.0000,0.0000,0.5535}{the}
\textcolor[rgb]{0.0000,0.0000,0.5535}{person}
\textcolor[rgb]{0.0000,0.0000,0.5713}{dressed}
\textcolor[rgb]{0.0000,0.0000,0.7852}{properly}
\textcolor[rgb]{0.0000,0.0000,1.0000}{for}
\textcolor[rgb]{0.0000,0.0490,1.0000}{the}
\textcolor[rgb]{0.0980,1.0000,0.8697}{sport}
\textcolor[rgb]{0.9278,0.0153,0.0000}{?}
}	
\end{center}   
\end{minipage}
\caption{Visualization of co-attention maps on success cases in the COCO-QA (first three columns using $\mathrm{Ours}^p$+VGG) and VQA (last two columns $\mathrm{Ours}^a$+VGG) dataset. The layout is the same as Fig.~\ref{fig:vis}.}
\label{fig:vis2}
\end{figure}

\begin{figure}[tp]
\begin{minipage}{0.19\linewidth}
\includegraphics[width=1\linewidth,  height=0.7\linewidth, scale=0.8, trim={0 0in 0 0in}, clip]{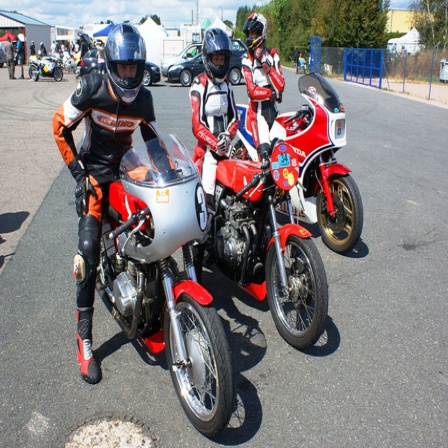}
\end{minipage}
\begin{minipage}{0.19\linewidth}
\includegraphics[width=1\linewidth,  height=0.7\linewidth, scale=0.8, trim={0 0in 0 0in}, clip]{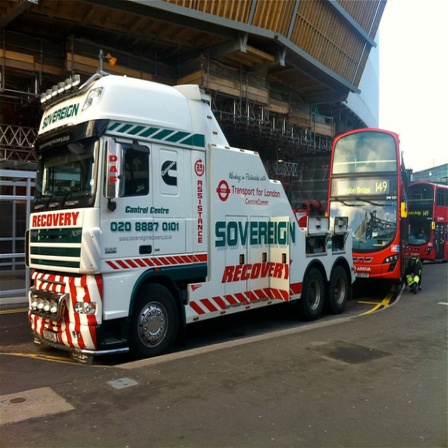}
\end{minipage}
\begin{minipage}{0.19\linewidth}
\includegraphics[width=1\linewidth,  height=0.7\linewidth, scale=0.8, trim={0 0in 0 0in}, clip]{demo/coco_qa/img_305.jpg}
\end{minipage}
\begin{minipage}{0.19\linewidth}
\includegraphics[width=1\linewidth,  height=0.7\linewidth, scale=0.8, trim={0 0in 0 0in}, clip]{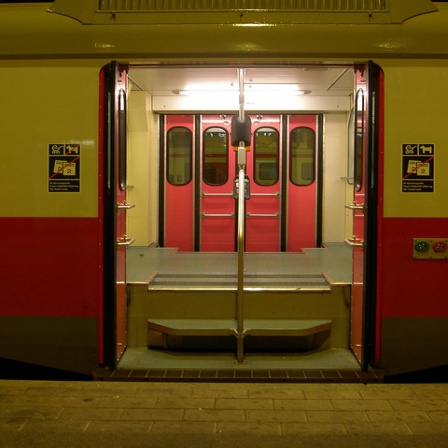}
\end{minipage}
\begin{minipage}{0.19\linewidth}
\includegraphics[width=1\linewidth,  height=0.7\linewidth, scale=0.8, trim={0 0in 0 0in}, clip]{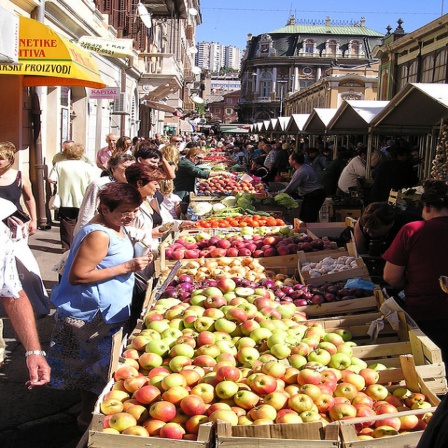}
\end{minipage}

\begin{minipage}{0.19\linewidth}
\begin{center}
\tiny{\textbf{Q}: how many red motorcycles with riders in protective gear are on the street? \textbf{A}: \textcolor{red}{\textbf{two}}(\textcolor{green}{\textbf{three}})}
\end{center}
\end{minipage}
\begin{minipage}{0.19\linewidth}
\begin{center}
\tiny{\textbf{Q}: what is the color of the bus? \textbf{A}: \textcolor{red}{\textbf{green}}(\textcolor{green}{\textbf{red}})}
\end{center}
\end{minipage}
\begin{minipage}{0.19\linewidth}
\begin{center}
\tiny{\textbf{Q}: what is shown in different places? \textbf{A}: \textcolor{red}{\textbf{hydrant}}(\textcolor{green}{\textbf{toy}})}
\end{center}
\end{minipage}
\begin{minipage}{0.19\linewidth}
\begin{center}
\tiny{\textbf{Q}:do the doors open in or out? \\ \textbf{A}: \textcolor{red}{\textbf{open}}(\textcolor{green}{\textbf{in}})}
\end{center}
\end{minipage}
\begin{minipage}{0.19\linewidth}
\begin{center}
\tiny{\textbf{Q}: is this an open market at night?  \textbf{A}: \textcolor{red}{\textbf{yes}}(\textcolor{green}{\textbf{no}})}
\end{center}   
\end{minipage}

\begin{minipage}{0.19\linewidth}
\includegraphics[width=1\linewidth,  height=0.7\linewidth, scale=0.8, trim={0 0in 0 0in}, clip]{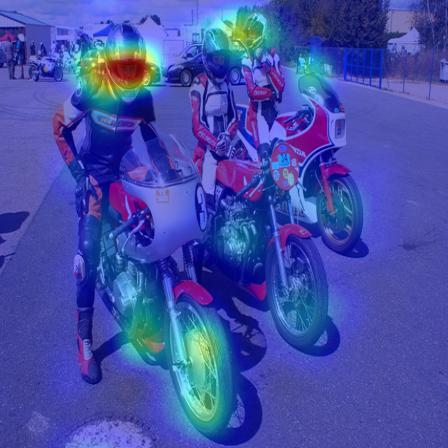}
\end{minipage}
\begin{minipage}{0.19\linewidth}
\includegraphics[width=1\linewidth,  height=0.7\linewidth, scale=0.8, trim={0 0in 0 0in}, clip]{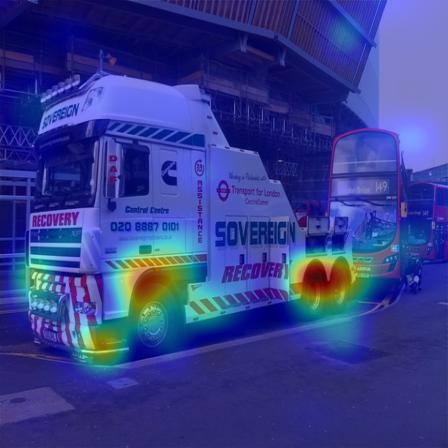}
\end{minipage}
\begin{minipage}{0.19\linewidth}
\includegraphics[width=1\linewidth,  height=0.7\linewidth, scale=0.8, trim={0 0in 0 0in}, clip]{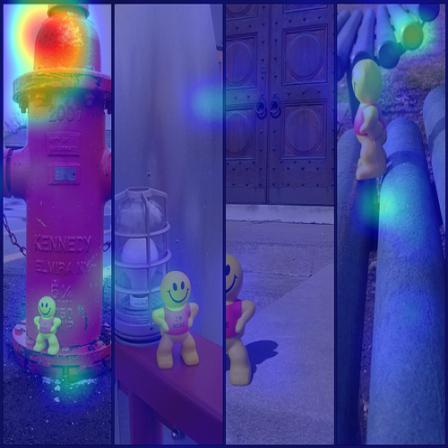}
\end{minipage}
\begin{minipage}{0.19\linewidth}
\includegraphics[width=1\linewidth,  height=0.7\linewidth, scale=0.8, trim={0 0in 0 0in}, clip]{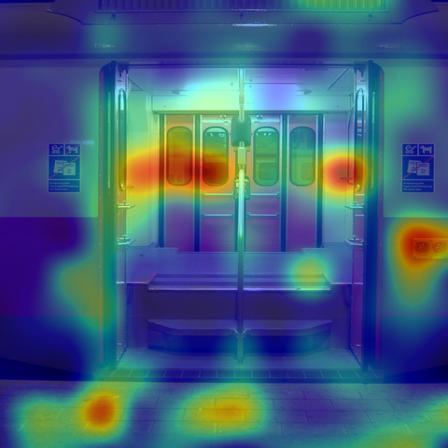}
\end{minipage}
\begin{minipage}{0.19\linewidth}
\includegraphics[width=1\linewidth,  height=0.7\linewidth, scale=0.8, trim={0 0in 0 0in}, clip]{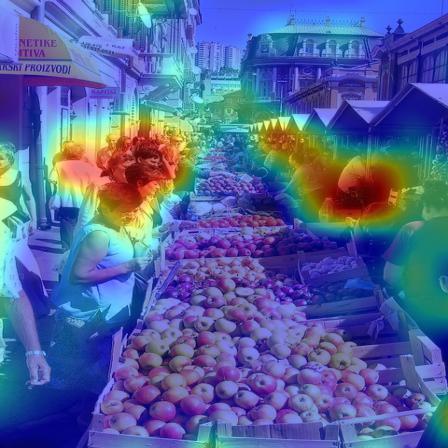}
\end{minipage}

\begin{minipage}{0.19\linewidth}
\begin{center}
\tiny{
\setlength{\fboxsep}{\fsize}
%\attention{0.063}{how}
%\attention{0.057}{many}
%\attention{0.071}{red}
%\attention{0.104}{motorcycles}
%\attention{0.056}{with}
%\attention{0.057}{riders}
%\attention{0.057}{in}
%\attention{0.071}{protective}
%\attention{0.098}{gear}
%\attention{0.059}{are}
%\attention{0.062}{on}
%\attention{0.060}{the}
%\attention{0.108}{street}
%\attention{0.077}{?}
\textcolor[rgb]{0.0000,0.4098,1.0000}{how}
\textcolor[rgb]{0.0000,0.3157,1.0000}{many}
\textcolor[rgb]{0.0000,0.5196,1.0000}{red}
\textcolor[rgb]{0.0854,1.0000,0.8824}{motorcycles}
\textcolor[rgb]{0.0000,0.3157,1.0000}{with}
\textcolor[rgb]{0.0000,0.3157,1.0000}{riders}
\textcolor[rgb]{0.0000,0.3157,1.0000}{in}
\textcolor[rgb]{0.0000,0.5196,1.0000}{protective}
\textcolor[rgb]{0.0221,0.9275,0.9456}{gear}
\textcolor[rgb]{0.0000,0.3471,1.0000}{are}
\textcolor[rgb]{0.0000,0.3941,1.0000}{on}
\textcolor[rgb]{0.0000,0.3627,1.0000}{the}
\textcolor[rgb]{0.1360,1.0000,0.8318}{street}
\textcolor[rgb]{0.0000,0.6137,1.0000}{?}
}
\end{center}
\end{minipage}
\begin{minipage}{0.19\linewidth}
\begin{center}
\tiny{
\setlength{\fboxsep}{\fsize}
%\attention{0.112}{what}
%\attention{0.137}{is}
%\attention{0.115}{the}
%\attention{0.127}{color}
%\attention{0.129}{of}
%\attention{0.115}{the}
%\attention{0.134}{bus}
%\attention{0.130}{?}
\textcolor[rgb]{0.0000,0.7549,1.0000}{what}
\textcolor[rgb]{0.1233,1.0000,0.8444}{is}
\textcolor[rgb]{0.0000,0.8020,1.0000}{the}
\textcolor[rgb]{0.0221,0.9275,0.9456}{color}
\textcolor[rgb]{0.0474,0.9588,0.9203}{of}
\textcolor[rgb]{0.0000,0.8020,1.0000}{the}
\textcolor[rgb]{0.0854,1.0000,0.8824}{bus}
\textcolor[rgb]{0.0474,0.9588,0.9203}{?}
}
\end{center}
\end{minipage}
\begin{minipage}{0.19\linewidth}
\begin{center}
\tiny{
\setlength{\fboxsep}{\fsize}
%\attention{0.105}{what}
%\attention{0.096}{is}
%\attention{0.096}{shown}
%\attention{0.097}{in}
%\attention{0.281}{different}
%\attention{0.230}{places}
%\attention{0.097}{?}
\textcolor[rgb]{0.0000,0.4882,1.0000}{what}
\textcolor[rgb]{0.0000,0.3941,1.0000}{is}
\textcolor[rgb]{0.0000,0.3941,1.0000}{shown}
\textcolor[rgb]{0.0000,0.4098,1.0000}{in}
\textcolor[rgb]{1.0000,0.9158,0.0000}{different}
\textcolor[rgb]{0.6167,1.0000,0.3510}{places}
\textcolor[rgb]{0.0000,0.4098,1.0000}{?}
}
\end{center}
\end{minipage}
\begin{minipage}{0.19\linewidth}
\begin{center}
\tiny{
\setlength{\fboxsep}{\fsize}
%\attention{0.231}{do}
%\attention{0.044}{the}
%\attention{0.251}{doors}
%\attention{0.230}{open}
%\attention{0.047}{in}
%\attention{0.063}{or}
%\attention{0.093}{out}
%\attention{0.041}{?}
\textcolor[rgb]{0.5914,1.0000,0.3763}{do}
\textcolor[rgb]{0.0000,0.0000,0.9456}{the}
\textcolor[rgb]{0.7432,1.0000,0.2245}{doors}
\textcolor[rgb]{0.5787,1.0000,0.3890}{open}
\textcolor[rgb]{0.0000,0.0000,0.9813}{in}
\textcolor[rgb]{0.0000,0.0804,1.0000}{or}
\textcolor[rgb]{0.0000,0.3471,1.0000}{out}
\textcolor[rgb]{0.0000,0.0000,0.9278}{?}
}
\end{center}
\end{minipage}
\begin{minipage}{0.19\linewidth}
\begin{center}
\tiny{
\setlength{\fboxsep}{\fsize}
%\attention{0.052}{is}
%\attention{0.102}{this}
%\attention{0.094}{an}
%\attention{0.245}{open}
%\attention{0.199}{market}
%\attention{0.134}{at}
%\attention{0.119}{night}
%\attention{0.055}{?}
\textcolor[rgb]{0.0000,0.0176,1.0000}{is}
\textcolor[rgb]{0.0000,0.5196,1.0000}{this}
\textcolor[rgb]{0.0000,0.4412,1.0000}{an}
\textcolor[rgb]{0.8697,1.0000,0.0980}{open}
\textcolor[rgb]{0.4902,1.0000,0.4775}{market}
\textcolor[rgb]{0.0000,0.8490,1.0000}{at}
\textcolor[rgb]{0.0000,0.6922,1.0000}{night}
\textcolor[rgb]{0.0000,0.0490,1.0000}{?}
}
\end{center}   
\end{minipage}

\begin{minipage}{0.19\linewidth}
\includegraphics[width=1\linewidth,  height=0.7\linewidth, scale=0.8, trim={0 0in 0 0in}, clip]{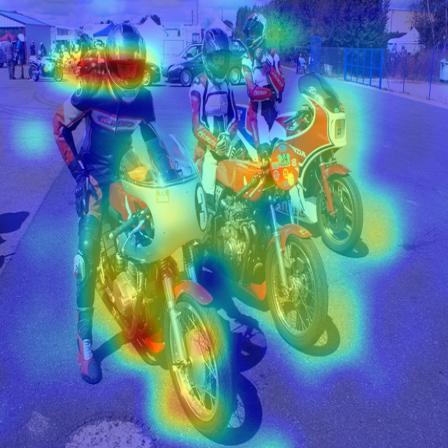}
\end{minipage}
\begin{minipage}{0.19\linewidth}
\includegraphics[width=1\linewidth,  height=0.7\linewidth, scale=0.8, trim={0 0in 0 0in}, clip]{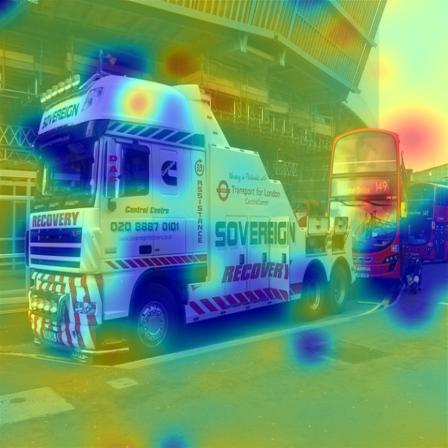}
\end{minipage}
\begin{minipage}{0.19\linewidth}
\includegraphics[width=1\linewidth,  height=0.7\linewidth, scale=0.8, trim={0 0in 0 0in}, clip]{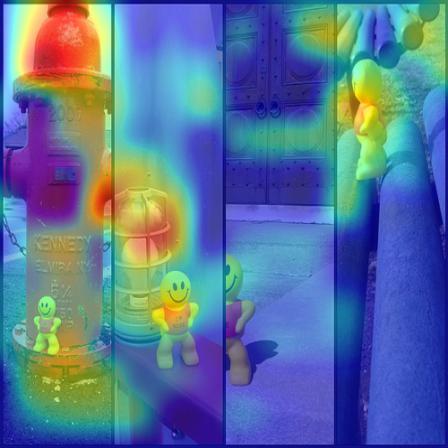}
\end{minipage}
\begin{minipage}{0.19\linewidth}
\includegraphics[width=1\linewidth,  height=0.7\linewidth, scale=0.8, trim={0 0in 0 0in}, clip]{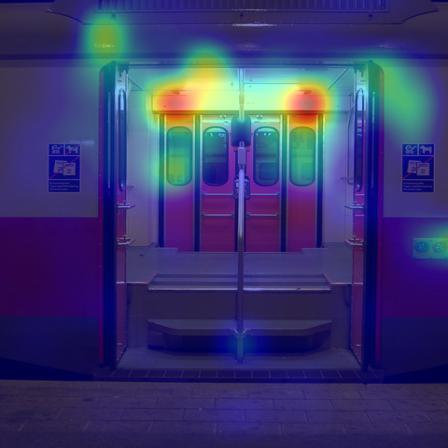}
\end{minipage}
\begin{minipage}{0.19\linewidth}
\includegraphics[width=1\linewidth,  height=0.7\linewidth, scale=0.8, trim={0 0in 0 0in}, clip]{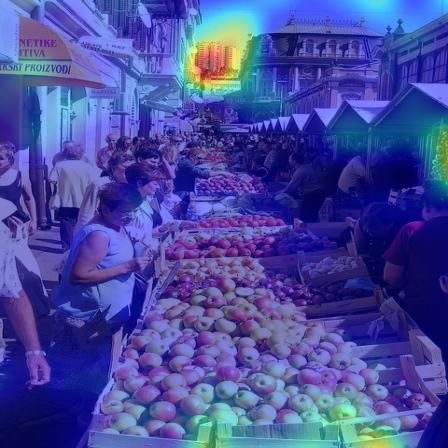}
\end{minipage}

\vspace{0pt}
\begin{minipage}{0.19\linewidth}
\begin{center}
\tiny{
\setlength{\fboxsep}{\fsize}
%\attention{0.074}{how}
%\attention{0.511}{many}
%\attention{0.038}{red}
%\attention{0.083}{motorcycles}
%\attention{0.025}{with}
%\attention{0.025}{riders}
%\attention{0.029}{in}
%\attention{0.039}{protective}
%\attention{0.044}{gear}
%\attention{0.025}{are}
%\attention{0.030}{on}
%\attention{0.029}{the}
%\attention{0.024}{street}
%\attention{0.024}{?}
\textcolor[rgb]{0.0000,0.0490,1.0000}{how}
\textcolor[rgb]{0.6783,0.0000,0.0000}{many}
\textcolor[rgb]{0.0000,0.0000,0.8209}{red}
\textcolor[rgb]{0.0000,0.1118,1.0000}{motorcycles}
\textcolor[rgb]{0.0000,0.0000,0.7139}{with}
\textcolor[rgb]{0.0000,0.0000,0.7139}{riders}
\textcolor[rgb]{0.0000,0.0000,0.7317}{in}
\textcolor[rgb]{0.0000,0.0000,0.8209}{protective}
\textcolor[rgb]{0.0000,0.0000,0.8743}{gear}
\textcolor[rgb]{0.0000,0.0000,0.7139}{are}
\textcolor[rgb]{0.0000,0.0000,0.7496}{on}
\textcolor[rgb]{0.0000,0.0000,0.7317}{the}
\textcolor[rgb]{0.0000,0.0000,0.6961}{street}
\textcolor[rgb]{0.0000,0.0000,0.6961}{?}
}
\end{center}
\end{minipage}
\begin{minipage}{0.19\linewidth}
\begin{center}
\tiny{
\setlength{\fboxsep}{\fsize}
%\attention{0.035}{what}
%\attention{0.037}{is}
%\attention{0.033}{the}
%\attention{0.758}{color}
%\attention{0.033}{of}
%\attention{0.034}{the}
%\attention{0.032}{bus}
%\attention{0.037}{?}
\textcolor[rgb]{0.0000,0.0000,0.6961}{what}
\textcolor[rgb]{0.0000,0.0000,0.7139}{is}
\textcolor[rgb]{0.0000,0.0000,0.6961}{the}
\textcolor[rgb]{0.5178,0.0000,0.0000}{color}
\textcolor[rgb]{0.0000,0.0000,0.6961}{of}
\textcolor[rgb]{0.0000,0.0000,0.6961}{the}
\textcolor[rgb]{0.0000,0.0000,0.6783}{bus}
\textcolor[rgb]{0.0000,0.0000,0.7139}{?}
}
\end{center}
\end{minipage}
\begin{minipage}{0.19\linewidth}
\begin{center}
\tiny{
\setlength{\fboxsep}{\fsize}
%\attention{0.078}{what}
%\attention{0.086}{is}
%\attention{0.086}{shown}
%\attention{0.103}{in}
%\attention{0.112}{different}
%\attention{0.432}{places}
%\attention{0.103}{?}
\textcolor[rgb]{0.0000,0.1275,1.0000}{what}
\textcolor[rgb]{0.0000,0.1902,1.0000}{is}
\textcolor[rgb]{0.0000,0.1902,1.0000}{shown}
\textcolor[rgb]{0.0000,0.3314,1.0000}{in}
\textcolor[rgb]{0.0000,0.4098,1.0000}{different}
\textcolor[rgb]{1.0000,0.1024,0.0000}{places}
\textcolor[rgb]{0.0000,0.3314,1.0000}{?}
}
\end{center}
\end{minipage}
\begin{minipage}{0.19\linewidth}
\begin{center}
\tiny{
\setlength{\fboxsep}{\fsize}
%\attention{0.055}{what}
%\attention{0.034}{the}
%\attention{0.069}{doors}
%\attention{0.128}{open}
%\attention{0.035}{in}
%\attention{0.570}{or}
%\attention{0.069}{out}
%\attention{0.039}{?}
\textcolor[rgb]{0.0000,0.0000,0.9100}{do}
\textcolor[rgb]{0.0000,0.0000,0.7496}{the}
\textcolor[rgb]{0.0000,0.0000,1.0000}{doors}
\textcolor[rgb]{0.0000,0.3471,1.0000}{open}
\textcolor[rgb]{0.0000,0.0000,0.7496}{in}
\textcolor[rgb]{0.7139,0.0000,0.0000}{or}
\textcolor[rgb]{0.0000,0.0000,1.0000}{out}
\textcolor[rgb]{0.0000,0.0000,0.7852}{?}
}
\end{center}
\end{minipage}
\begin{minipage}{0.19\linewidth}
\begin{center}
\tiny{
\setlength{\fboxsep}{\fsize}
%\attention{0.054}{is}
%\attention{0.069}{this}
%\attention{0.089}{an}
%\attention{0.174}{open}
%\attention{0.302}{market}
%\attention{0.080}{at}
%\attention{0.174}{night}
%\attention{0.057}{?}
\textcolor[rgb]{0.0000,0.0020,1.0000}{is}
\textcolor[rgb]{0.0000,0.1431,1.0000}{this}
\textcolor[rgb]{0.0000,0.3471,1.0000}{an}
\textcolor[rgb]{0.1992,1.0000,0.7685}{open}
\textcolor[rgb]{1.0000,0.7124,0.0000}{market}
\textcolor[rgb]{0.0000,0.2529,1.0000}{at}
\textcolor[rgb]{0.1992,1.0000,0.7685}{night}
\textcolor[rgb]{0.0000,0.0333,1.0000}{?}
}
\end{center}   
\end{minipage}

\begin{minipage}{0.19\linewidth}
\includegraphics[width=1\linewidth,  height=0.7\linewidth, scale=0.8, trim={0 0in 0 0in}, clip]{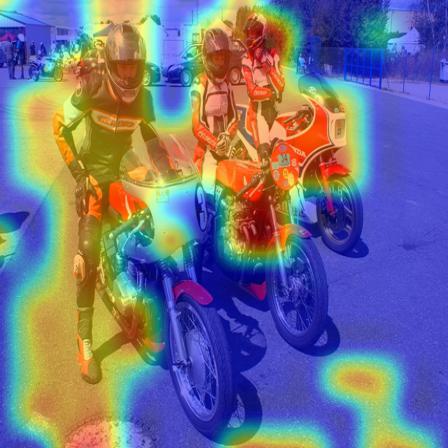}
\end{minipage}
\begin{minipage}{0.19\linewidth}
\includegraphics[width=1\linewidth,  height=0.7\linewidth, scale=0.8, trim={0 0in 0 0in}, clip]{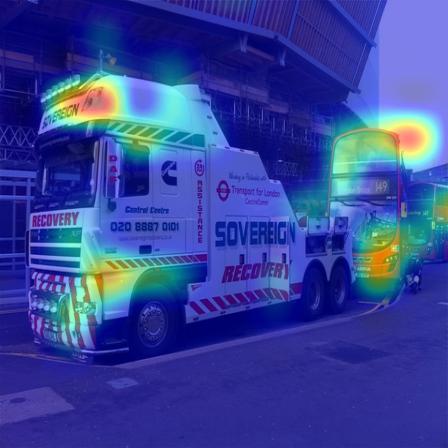}
\end{minipage}
\begin{minipage}{0.19\linewidth}
\includegraphics[width=1\linewidth,  height=0.7\linewidth, scale=0.8, trim={0 0in 0 0in}, clip]{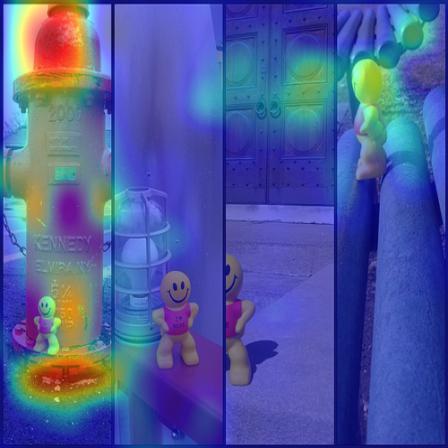}
\end{minipage}
\begin{minipage}{0.19\linewidth}
\includegraphics[width=1\linewidth,  height=0.7\linewidth, scale=0.8, trim={0 0in 0 0in}, clip]{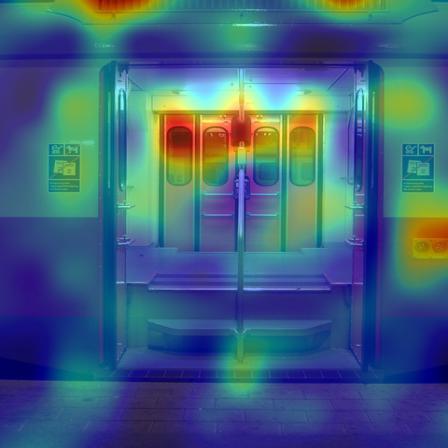}
\end{minipage}
\begin{minipage}{0.19\linewidth}
\includegraphics[width=1\linewidth,  height=0.7\linewidth, scale=0.8, trim={0 0in 0 0in}, clip]{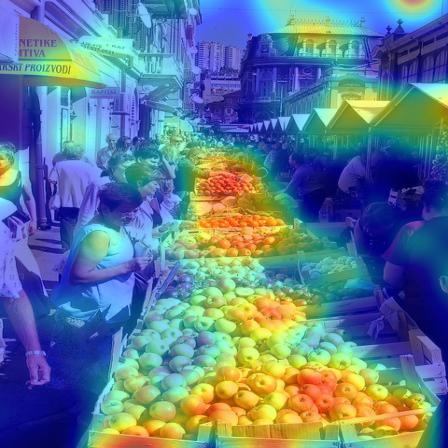}
\end{minipage}

\vspace{0pt}
\begin{minipage}{0.19\linewidth}
\begin{center}
\tiny{
\setlength{\fboxsep}{\fsize}
%\attention{0.056}{how}
%\attention{0.055}{many}
%\attention{0.060}{red}
%\attention{0.064}{motorcycles}
%\attention{0.071}{with}
%\attention{0.066}{riders}
%\attention{0.072}{in}
%\attention{0.070}{protective}
%\attention{0.067}{gear}
%\attention{0.067}{are}
%\attention{0.067}{on}
%\attention{0.088}{the}
%\attention{0.089}{street}
%\attention{0.105}{?}
\textcolor[rgb]{0.0000,0.3157,1.0000}{how}
\textcolor[rgb]{0.0000,0.3000,1.0000}{many}
\textcolor[rgb]{0.0000,0.3784,1.0000}{red}
\textcolor[rgb]{0.0000,0.4412,1.0000}{motorcycles}
\textcolor[rgb]{0.0000,0.5510,1.0000}{with}
\textcolor[rgb]{0.0000,0.4725,1.0000}{riders}
\textcolor[rgb]{0.0000,0.5510,1.0000}{in}
\textcolor[rgb]{0.0000,0.5353,1.0000}{protective}
\textcolor[rgb]{0.0000,0.4882,1.0000}{gear}
\textcolor[rgb]{0.0000,0.4882,1.0000}{are}
\textcolor[rgb]{0.0000,0.4882,1.0000}{on}
\textcolor[rgb]{0.0000,0.8020,1.0000}{the}
\textcolor[rgb]{0.0000,0.8176,1.0000}{street}
\textcolor[rgb]{0.1233,1.0000,0.8444}{?}
}
\end{center}
\end{minipage}
\begin{minipage}{0.19\linewidth}
\begin{center}
\tiny{
\setlength{\fboxsep}{\fsize}
%\attention{0.053}{what}
%\attention{0.049}{is}
%\attention{0.050}{the}
%\attention{0.180}{color}
%\attention{0.181}{of}
%\attention{0.166}{the}
%\attention{0.172}{bus}
%\attention{0.150}{?}
\textcolor[rgb]{0.0000,0.0333,1.0000}{what}
\textcolor[rgb]{0.0000,0.0020,1.0000}{is}
\textcolor[rgb]{0.0000,0.0020,1.0000}{the}
\textcolor[rgb]{0.3510,1.0000,0.6167}{color}
\textcolor[rgb]{0.3637,1.0000,0.6040}{of}
\textcolor[rgb]{0.2372,1.0000,0.7306}{the}
\textcolor[rgb]{0.2878,1.0000,0.6799}{bus}
\textcolor[rgb]{0.1107,1.0000,0.8571}{?}
}
\end{center}
\end{minipage}
\begin{minipage}{0.19\linewidth}
\begin{center}
\tiny{
\setlength{\fboxsep}{\fsize}
%\attention{0.103}{what}
%\attention{0.106}{is}
%\attention{0.106}{shown}
%\attention{0.176}{in}
%\attention{0.181}{different}
%\attention{0.165}{places}
%\attention{0.163}{?}
\textcolor[rgb]{0.0000,0.5510,1.0000}{what}
\textcolor[rgb]{0.0000,0.5824,1.0000}{is}
\textcolor[rgb]{0.0000,0.5824,1.0000}{shown}
\textcolor[rgb]{0.3384,1.0000,0.6293}{in}
\textcolor[rgb]{0.3763,1.0000,0.5914}{different}
\textcolor[rgb]{0.2372,1.0000,0.7306}{places}
\textcolor[rgb]{0.2245,1.0000,0.7432}{?}
}
\end{center}
\end{minipage}
\begin{minipage}{0.19\linewidth}
\begin{center}
\tiny{
\setlength{\fboxsep}{\fsize}
%\attention{0.006}{do}
%\attention{0.004}{the}
%\attention{0.004}{doors}
%\attention{0.007}{open}
%\attention{0.011}{in}
%\attention{0.131}{or}
%\attention{0.408}{out}
%\attention{0.408}{?}
\textcolor[rgb]{0.0000,0.0000,0.5357}{do}
\textcolor[rgb]{0.0000,0.0000,0.5178}{the}
\textcolor[rgb]{0.0000,0.0000,0.5178}{doors}
\textcolor[rgb]{0.0000,0.0000,0.5535}{open}
\textcolor[rgb]{0.0000,0.0000,0.5713}{in}
\textcolor[rgb]{0.0000,0.3784,1.0000}{or}
\textcolor[rgb]{1.0000,0.8141,0.0000}{out}
\textcolor[rgb]{1.0000,0.8141,0.0000}{?}
}
\end{center}
\end{minipage}
\begin{minipage}{0.19\linewidth}
\begin{center}
\tiny{
\setlength{\fboxsep}{\fsize}
%\attention{0.010}{is}
%\attention{0.010}{this}
%\attention{0.012}{an}
%\attention{0.023}{open}
%\attention{0.099}{market}
%\attention{0.183}{at}
%\attention{0.290}{night}
%\attention{0.373}{?}
\textcolor[rgb]{0.0000,0.0000,0.5713}{is}
\textcolor[rgb]{0.0000,0.0000,0.5713}{this}
\textcolor[rgb]{0.0000,0.0000,0.5891}{an}
\textcolor[rgb]{0.0000,0.0000,0.6961}{open}
\textcolor[rgb]{0.0000,0.2686,1.0000}{market}
\textcolor[rgb]{0.0095,0.9118,0.9583}{at}
\textcolor[rgb]{0.6799,1.0000,0.2878}{night}
\textcolor[rgb]{1.0000,0.6979,0.0000}{?}
}
\end{center}   
\end{minipage}
\caption{Visualization of co-attention maps on failure cases in the COCO-QA (first three columns using $\mathrm{Ours}^p$+VGG) and VQA (last two columns $\mathrm{Ours}^a$+VGG) dataset. Predicted answer is in red and ground truth answer is in green. The layout is the same as Fig.~\ref{fig:vis}.}
\label{fig:vis3}
\end{figure}
%%%%%%%%%%%%%%%%%%%%%%%%%%%%%%%%%%%%%%%%%%%%%%%%%%%%%%%%%%
% References
%%%%%%%%%%%%%%%%%%%%%%%%%%%%%%%%%%%%%%%%%%%%%%%%%%%%%%%%%%%
{\small
\bibliographystyle{plain}
\bibliography{egref}
}
\end{document}